\let\mypdfximage\pdfximage
\def\pdfximage{\immediate\mypdfximage}

\documentclass[final]{cvpr}

\usepackage{times}
\usepackage{epsfig}
\usepackage{graphicx}
\usepackage{amsmath}
\usepackage{amssymb}

\usepackage{epstopdf}
\usepackage{bm}
\usepackage{multirow}

\usepackage{xspace}
\usepackage{float}
\usepackage{xcolor}
\usepackage{algorithm}
\usepackage{mathtools,xparse}
\usepackage{siunitx}
\usepackage{stmaryrd}
\usepackage{algpseudocode}
\usepackage{placeins}

\makeatletter
\@namedef{ver@everyshi.sty}{}
\makeatother
\usepackage{tikz}

\usepackage{ctable}
\usepackage{booktabs}

\usetikzlibrary{shapes.geometric}
\usetikzlibrary{shapes,arrows}

\usepackage{collcell}
\usepackage{xstring}
\usepackage{pgf}
\usepackage{adjustbox}
\usepackage{colortbl}

\usepackage{enumitem}

\usepackage{pifont}
\newcommand{\xmark}{\ding{55}}

\usepackage{pgfplotstable}
\usepgfplotslibrary{colormaps}

\newcommand{\boldparagraph}[1]{\vspace{0.01em}\noindent{\bf #1} }

\newcommand{\winner}[1]{\textcolor{black}{\textbf{#1}}}
\newcommand{\oracle}[1]{\textcolor{gray}{\textbf{#1}}}
\newcommand{\La}{\mathcal{L}}
\newcommand{\Lp}{\alpha}
\newcommand{\Lfm}{F}
\newcommand{\Li}{I}
\newcommand{\Ll}{L}
\newcommand{\Ls}{S}
\newcommand{\Lt}{T}
\newcommand{\LossWithPar}[1]{\Lp_#1 \La_#1}

\newcommand{\TI}{t}
\newcommand{\Ren}{R}
\newcommand{\Fen}{F}
\newcommand{\Men}{M}

\newcommand{\Fi}{\Fen_\TI}
\newcommand{\Mi}{\Men_\TI}
\newcommand{\Ri}{\Ren_\TI}

\newcommand{\GTS}[1]{\tilde{#1}}
\newcommand{\GTFi}{\GTS{\Fen}_\TI}
\newcommand{\GTMi}{\GTS{\Men}_\TI}
\newcommand{\GTRi}{\GTS{\Ren}_\TI}

\newcommand{\expos}{\epsilon}

\newcommand{\MySet}[1]{\{#1\}_0^1}

\newcommand\blfootnote[1]{%
  \begingroup
  \renewcommand\thefootnote{}\footnote{#1}%
  \addtocounter{footnote}{-1}%
  \endgroup
}

\newcommand{\mathfont}{\small}

\usepackage[pagebackref=true,breaklinks=true,colorlinks,bookmarks=false]{hyperref}

\def\cvprPaperID{548} 
\def\confYear{CVPR 2021}
\begin{document}

\title{DeFMO: Deblurring and Shape Recovery of Fast Moving Objects}


\author{Denys Rozumnyi$^{1,4}$
\hspace{0.5cm}
Martin R. Oswald$^{1}$
\hspace{0.5cm}
Vittorio Ferrari$^{2}$
\hspace{0.5cm}
Ji\v{r}\'\i{} Matas$^{4}$
\hspace{0.5cm}
Marc Pollefeys$^{1,3}$\\[1.2em]
{\normalsize $^{1}$Department of Computer Science, ETH Zurich }
\hspace{1.5cm} 
{\normalsize  $^{3}$Microsoft Mixed Reality and AI Zurich Lab } \\
{\normalsize $^{2}$Google Research}
\hspace{1.5cm}
{\normalsize $^{4}$Visual Recognition Group, Czech Technical University in Prague}
}

\maketitle


\begin{abstract}
Objects moving at high speed appear significantly blurred when captured with cameras. The blurry appearance is especially ambiguous when the object has complex shape or texture. In such cases, classical methods, or even humans, are unable to recover the object's appearance and motion. We propose a method that, given a single image with its estimated background, outputs the object's appearance and position in a series of sub-frames as if captured by a high-speed camera (i.e.~temporal super-resolution). The proposed generative model embeds an image of the blurred object into a latent space representation, disentangles the background, and renders the sharp appearance. Inspired by the image formation model, we design novel self-supervised loss function terms that boost performance and show good generalization capabilities. The proposed DeFMO method is trained on a complex synthetic dataset, yet it performs well on real-world data from several datasets. DeFMO outperforms the state of the art and generates high-quality temporal super-resolution frames. 
\end{abstract}

\section{Introduction}

Object blurring is a challenging problem in many image processing and computer vision tasks. 
The primary sources of image blur are rapid camera motion and object motion combined with long exposure time.
Many methods were proposed to address the deblurring task, ranging from image deblurring~\cite{Kupyn_2019_ICCV,Zhang_2020_CVPR} to video temporal super-resolution~\cite{Niklaus_CVPR_2020, Shen_2020_CVPR, xiang2020zooming}.
However, they consider only low to medium blur, emerging from global camera blur due to camera motion, defocused camera, or objects moving at moderate speed.  

\newcommand{\addimg}[1]{\includegraphics[width=0.082\linewidth]{imgs/teaser/#1}}
\newcommand{\drawArrow}[1]{ \multirow{3}{*}{\raisebox{0.0\linewidth}{\begin{tikzpicture}
  \draw[->, ultra thick] (0,0) -- node [above] {\rotatebox[origin=l]{270}{\tiny #1}}  (0.3,0);
  \draw[->, ultra thick] (0,0.85) -- (0.3,0.85);
\end{tikzpicture}}}}
\newcommand{\drawArrowNew}{\multirow{4}{*}{\large $\Rightarrow$}}
\newcommand{\makeRowT}[4]{
	\multirow{1}{*}[#3]{{\rotatebox{90}{\tiny #2}}} &
	\addimg{#1_bgr}&\addimg{#1_im}& #4 &\addimg{#1_est0}&\addimg{#1_est1}&\addimg{#1_est2}&\addimg{#1_est3}&\addimg{#1_est4}&\addimg{#1_est5}&\addimg{#1_est6}&\addimg{#1_est7}&\addimg{#1_hs7}\\}
\begin{figure}
\centering
\small
\setlength{\tabcolsep}{0.02em} 
\renewcommand{\arraystretch}{0.15} 

\begin{tabular}{@{}lccccccccccc@{\hskip 0.3em}c@{}}

\makeRowT{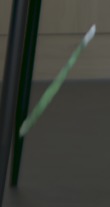}{pen}{21pt}{}
\makeRowT{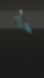}{key}{20pt}{}
\makeRowT{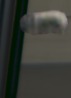}{eraser}{22pt}{} 
\makeRowT{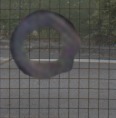}{aerobie}{16pt}{\drawArrowNew}
\makeRowT{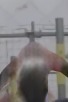}{volleyball}{20pt}{}
\makeRowT{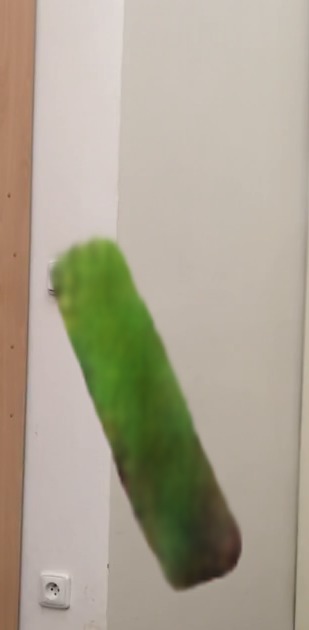}{lighter}{27pt}{}
\makeRowT{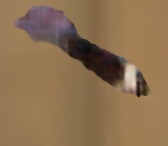}{mic}{11pt}{}
\makeRowT{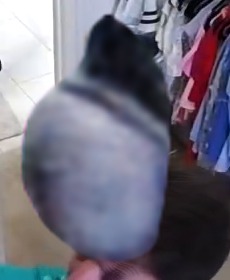}{cap}{15pt}{}

\\[0.2em]
& $B$ & $I$ &  & $I_{0/7}$ & $I_{1/7}$ & $I_{2/7}$ & $I_{3/7}$ & $I_{4/7}$ & $I_{5/7}$ & $I_{6/7}$ & $I_{7/7}$  & GT \\[0.7em]

\end{tabular}
\caption{\textbf{Temporal super-resolution.} 
Given an input image~$I$ depicting a blurred fast moving object and an estimated background~$B$, DeFMO decomposes the image into a series of deblurred sub-frames with sharp object contours.
Examples are from test datasets~\cite{kotera2020,tbd,tbd3d}, mobile device footage (lighter), and YouTube videos (mic, cap). 
Ground truth (GT) corresponds either to the high-speed camera frame or a static image. 
Deblurred images $I_t$ are sharper than the 'GT',~\eg 'pen' or 'key'. 
}
\label{fig:teaser}
\end{figure}

Only recently, specialized algorithms for the deblurring of fast moving objects (FMOs) have been introduced~\cite{kotera2020,tbd,tbd3d}. 
FMOs are defined as objects that move over a distance larger than their size within the camera exposure time (or within a single time frame in video).
FMO detection is important in tracking sports with fast object motion like soccer, tennis, or badminton. 
It is also beneficial in autonomous driving to detect impacts with stones, birds, or other wildlife.
FMOs are frequently found when capturing falling or thrown objects like pieces of fruit, leaves, flying insects, hailstorm, or rain.
Any moving object becomes an FMO in low-light conditions or for long exposure.
In other words, one needs to increase the exposure time or the speed of the object to observe an FMO.

We consider a setting where the input is an image with an object moving fast and thus appearing blurred. 
The task is to reconstruct the hypothetical sub-frames that would have been there if this was a short video captured by a high-speed camera for the same time interval. 
The physical generative model that leads to the input blurred frame is assumed to be a temporal integration of underlying sharp sub-frames, each of which has a much shorter exposure time. 
To simplify the problem complexity, we assume that background without the object is given,~\eg from previous frames in the video or as a static image captured when there is no object. 
In practice, a median of several previous frames works well.

Prior work on FMO deblurring considers only relatively simple, mostly spherical objects~\cite{tbd,fmo,tbd3d}.
This prior work typically assumes that the object in motion has a constant appearance in all sub-frames. 

We propose DeFMO~--~the first to go beyond these assumptions by handling the time-varying complex appearance of fast moving objects that move over 3D trajectories with 3D rotation.
DeFMO is a generative model that reconstructs sharp contours and appearance of FMOs.
First, we disentangle the blurred fast moving object from the background into a latent space. 
Then, a rendering network has the objective to render the sharp object in a series of sub-frames, capturing the motion in time.
The network is trained end-to-end on a synthetic dataset with complex, highly textured objects. 
Thanks to self-supervised loss function terms inspired by the image formation model with FMOs, our method easily generalizes to real-world data, as shown in Fig.~\ref{fig:teaser}.
DeFMO can be applied to many fields, such as video temporal super-resolution, data compression, surveillance, astronomy, and microscopy. 
Overall, the paper makes the following \textbf{contributions}:
\begin{itemize}[itemsep=0.1pt,topsep=3pt,leftmargin=*]
    \item We present the first fully neural network model for FMO deblurring that bridges the gap between deblurring, 3D modeling, and sub-frame tracking of FMOs. 
    \item Training only on synthetic data with novel self-supervised losses sets a new state of the art in terms of trajectory and sharp appearance reconstruction of FMOs. 
    \item We introduce a new synthetic dataset with complex objects, textures, and backgrounds. The dataset and model implementation are made publicly available\footnote{\url{https://github.com/rozumden/DeFMO}}.
\end{itemize}

\section{Related work}
\label{sec:rel-work}

\boldparagraph{Fast moving objects} were defined in~\cite{fmo}, and a proof-of-concept method was designed.
The blurring and matting (blatting) equation was later introduced in~\cite{tbd,kotera2018} as
\begin{equation}
\mathfont
	\label{eq:blatting}
	I = H*F + (1-H*M)\,B \enspace,
\end{equation}
where the sharp object given by foreground appearance $F$ and segmentation mask $M$ is blurred and combined with the background $B$. The blur kernel $H$ identifies the object's trajectory such that $\| H \| = 1$.
An additional constraint is $F \leq M$,~\ie 0s in $M$ (background) imply 0s in $F$.

The blatting equation~\eqref{eq:blatting} assumes that the object appearance $F$ is static within one video frame $I$.
In contrast to our approach, the object must also travel in a 2D plane parallel to the camera plane.
Kotera~\etal~\cite{kotera2020} study the special case that a planar FMO only rotates within a 2D plane parallel to the camera plane.
An improved motion blur prior for FMOs was proposed in~\cite{sroubek2020}. 
Learned FMO detection is studied in ~\cite{fmodetect}.
Some of the limiting assumptions of TbD were lifted by the TbD-3D~\cite{tbd3d} method that assumed a piece-wise constant appearance as 
\begin{equation}
\mathfont
	\label{eq:blatting_pw}
	I = \sum_i H_i*F_i + \big(1-\sum_i H_i*M_i\big)\,B \enspace,
\end{equation}
where index $i$ corresponds to one part of the full trajectory $H = \sum_i H_i$ traveled within one frame $I$. 
Sub-frame appearances $F_i$ and masks $M_i$ account for a potentially rotating or deforming object.
Solving~\eqref{eq:blatting_pw} simultaneously for $H_i, F_i, M_i$ is in practice computationally infeasible.
To address this problem, TbD-3D solves~\eqref{eq:blatting_pw} for $F_i$, $M_i$ by initially assuming a simpler static appearance model~\eqref{eq:blatting} to estimate the trajectory $H$, then splitting the trajectory into the desired number of pieces, and finally estimating the sub-frame appearances in a second pass while keeping the trajectory fixed. 
We claim that solving~\eqref{eq:blatting_pw} in such a fashion is sub-optimal and similar to a chicken-and-egg problem since robust trajectory estimation requires to model the time-varying appearance and vice versa.
Hence, TbD-3D strongly depends on the initial trajectory estimation from an external module, such as TbD~\cite{tbd} or TbD-NC~\cite{tbdnc}.
In practice, TbD-3D only works when the object has a trivial shape (\eg a sphere) or appearance (\eg uniform color).

Methods, such as~\cite{purohit2019bringing} or~\cite{Jin_2018_CVPR}, have been proposed to generate a video from a single image, but fast motion is not considered.
Blurry video frame interpolation~\cite{Jin_2019_CVPR,Shen_2020_CVPR}, softmax splatting~\cite{Niklaus_CVPR_2020}, or zooming slow-motion~\cite{xiang2020zooming} are designed for video frame interpolation to increase the frame rate from several blurred inputs. Still, the considered motion blurs are small compared to what is caused by FMOs.

\begin{figure*}
\centering
\setlength{\tabcolsep}{0.03em} 
\renewcommand{\arraystretch}{0.15} 
\includegraphics[width=1.0\linewidth]{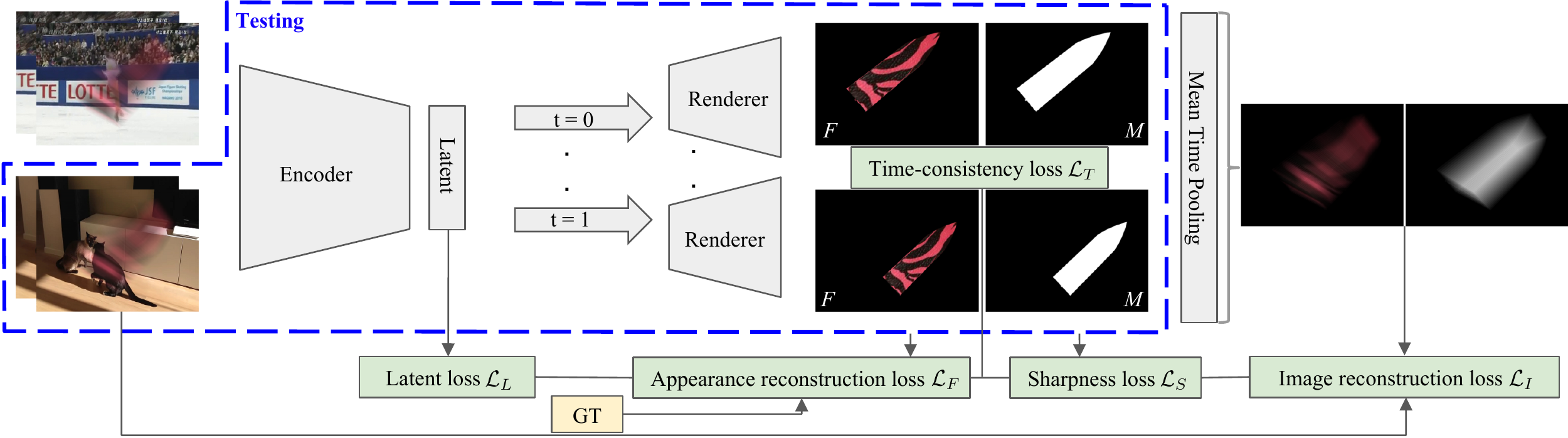}
\caption{\textbf{Architecture of DeFMO.} 
The input image and the estimated background are encoded into latent space $X$. 
Then, $X$ is augmented with time index channel $t$ and is rendered into the deblurred object with appearance $F$ and mask $M$. The renderings are generated for many time indices simultaneously with the same rendering network and are averaged in time. 
We only use the blue part during testing.
}
\label{fig:architecture}
\end{figure*}

\boldparagraph{Classical deblurring methods} have shown success, such as DeblurGAN-v2~\cite{Kupyn_2019_ICCV}, deblurring by realistic blurring~\cite{Zhang_2020_CVPR}, deblurring using deep priors~\cite{Ren_2020_CVPR}, and others~\cite{Nah,Pan_2019_CVPR}.
These methods assume that anything in the scene can be blurred.  
In our task, we assume a single object blurred due to its fast motion. 
Moreover, these methods generate only one deblurred frame, which is not sufficient for FMOs, where the desired result is a set of deblurred high-speed sub-frames.
In the experiments, we compare to DeblurGAN-v2~\cite{Kupyn_2019_ICCV} and Jin~\etal~\cite{Jin_2018_CVPR}.
As these methods are more general, they perform worse than ours on the specific case of blur caused by FMOs, on which we specialize.

In sum, all existing deblurring methods either lack proper modeling of FMOs or strongly rely on handcrafted priors and impose strong assumptions on the object's shape and appearance. 
Furthermore, current FMO deblurring methods are slow and take seconds per frame. They also rely on other external modules.  
To address these problems, our method estimates complex shape and appearance of FMOs, all in one network, and runs in real-time. 

\section{Deblurring model}
\label{sec:model}

The input to the method is an RGB image $I: D \subset \mathbb{R}^2 \to \mathbb{R}^3$ containing a blurred FMO and an estimate of the background $B: D \to \mathbb{R}^3$, which does not include the object of interest.
In most scenarios, a video stream is available, and $B$ can be estimated as a median of several previous frames, as such an operation will remove all FMOs~\cite{tbd}.


The desired output is a sharp rendering of the FMO for all sub-frames at predefined sub-frame time indices $\TI \in [0,1]$ for which we estimate 4-channel RGBA renderings $\Ri: D \to \mathbb{R}^4$.
These renderings are composed of an RGB part $\Fi: D \to \mathbb{R}^3$
(sharp appearance of the FMO)
and an alpha matting mask $\Mi: D \to \mathbb{R}$
(segmentation into foreground FMO and background,
Fig.~\ref{fig:architecture}). 

We encode the input image $I$ and background $B$ into a latent space $X \in  \mathbb{R}^K$. Then, we render a set of sub-frame appearances, which are pushed to be sharp, time-consistent, independent of the background, and which reconstruct the input image with the following image formation model
\begin{equation}
\mathfont
	\label{eq:blatting_int}
	I_{t_0:t_1} = \int_{t_0}^{t_1} \Fi \Mi \:\text{d}\TI + \Big(1-\int_{t_0}^{t_1} \Mi \:\text{d}\TI \Big)\,B \enspace,
\end{equation}
which is a generalization of the piecewise-constant FMO formation model~\eqref{eq:blatting_pw}. 
Instead of~\eqref{eq:blatting} and~\eqref{eq:blatting_pw}, we render the object directly at the desired location. 
We do not disentangle the trajectory blur kernels $H_t$, which are just Dirac deltas in our generalization~\eqref{eq:blatting_int}. The multiplication of appearance $\Fi$ by mask $\Mi$ replaces the constraint $\Fi \leq \Mi$,~\eg predicting low values in the mask will imply low values in $\Fi \Mi$.
We assume that the input image $I = I_{0:1}$.
For training, we partition the time duration of the input frame into $N$ equidistant parts to generate the sub-frames, and evaluate $\TI$ on the set $\{\frac{i-1}{N-1}\}_{i=1}^{N}$. For testing, any $\TI \in [0,1]$ produces consistent renderings.

As a technicality, we render images as they were captured with nearly zero exposure time. 
Such an image at sub-frame time index $t$ is captured as $I_{t} = \lim_{\delta \to 0} I_{t-\delta:t+\delta} = \Fi \Mi + (1 - \Mi) B$. 
In practice, we deliberately add time blur for quantitative evaluation to match the high-speed camera frames (see evaluation Sec.~\ref{sec:eval}).
Such time blur is generated as temporal super-resolution by $l$, $\{I_{k/l:(k+\expos)/l}\}_{k=0}^{l-1}$, where $\expos$ is the exposure fraction. 
The ability to generate zero-exposure-time images enables the creation of \textit{any} frame rates with \textit{any} exposure time.

\subsection{Training loss}
The loss function is inspired by the nature of the problem.
The main goal is to get sharp ($\La_\Ls$) reconstructions of the object ($\La_\Lfm$) that do not contain the background ($\La_\Ll$), move smoothly over time ($\La_\Lt$), and reconstruct the input image ($\La_\Li$) according to the generalized formation model~\eqref{eq:blatting_int}. 
Therefore, the loss is a combination of five parts,
\begin{equation}
\mathfont
	\label{eq:loss}
	\La = \La_\Lfm + \LossWithPar{\Li} + \LossWithPar{\Lt}  + \LossWithPar{\Ls} + \LossWithPar{\Ll} \enspace.
\end{equation}

\boldparagraph{Appearance reconstruction loss $\La_\Lfm$}
captures the supervised sub-frame reconstruction of the object's appearance and mask.
Since the input is a single blurred image, we do not know whether time goes forward or backward. 
In fact, both of them will generate the same blurred image. These two cases are indistinguishable, as~\eg in determining temporal order of images from 3D structure~\cite{4datlanta}. 
Therefore, we evaluate both directions and calculate the loss only with respect to the time direction that best aligns with the ground truth.
More precisely, we define $\La_\Lfm$ as
\begin{equation}
\mathfont
	\label{eq:lfm}
	\begin{split}
	\La_\Lfm \big(\MySet{\Ri,\GTRi}\big) = 
	\min \Big( \int_{0}^{1} \La_R(\Ri,\GTRi) \:\text{d}\TI, \\
	\int_{0}^{1} \La_R(\Ri,\GTS{\Ren}_{1-\TI}) \:\text{d}\TI \Big) \enspace,
	\end{split}
\end{equation}
and the rendering loss $\La_R$ for a pair of the estimated rendering $\Ri$ and the ground truth rendering $\GTRi$ as
\begin{equation}
\mathfont
	\label{eq:lr}
	\begin{split}
	&\La_R \Big(\Ri=(\Fi,\Mi),\GTRi=(\GTFi,\GTMi)\Big) = \La_1(\Mi, \GTMi, \GTMi > 0)  \\
	&  + \La_1(\Mi, \GTMi, \GTMi = 0) + \La_1(\Fi \Mi, \GTFi \GTMi, \GTMi > 0) \enspace,
	\end{split}
\end{equation}
where we combine three terms.
The first two terms evaluate the difference between masks on two sets of pixels: where the object is present, and the rest of the image, as given by the ground truth.
The last term calculates the difference between the appearances, evaluated on the first set of pixels.
The intuition of this splitting is that the object usually occupies only a fraction of the input image, and we want the loss to focus more on the object itself.
Then, the $\La_1$ loss is
\begin{equation}
\mathfont
	\label{eq:l1}
	\La_1(M_1,M_2,O) = \frac{1}{|O|} \sum_{p \in D} \|M_1(p) - M_2(p)\|_1 O(p) \enspace,
\end{equation}
where $M(p)$ denotes the value of the image $M$ at pixel location $p$. If the occupancy mask $O$ is omitted, the whole image domain $D$ is assumed.
The appearance reconstruction loss is the only one that requires ground truth object renderings $\GTRi$,~\ie it is the only supervised loss.

\boldparagraph{Image reconstruction loss $\La_\Li$} is a direct application of the generalized formation model~\eqref{eq:blatting_int}.
We penalize for the difference between the input image and the synthetic image reconstructed from the input background and FMO renderings. The image reconstruction loss is defined as
\begin{equation}
\mathfont
	\label{eq:lim}
	\La_\Li(\MySet{\Fi,\Mi}) = \La_1 \Big(I, \!\int_{0}^{1} \!\! \Fi \Mi \:\text{d}\TI + \big(1- \!\!\int_{0}^{1} \!\!\Mi \:\text{d}\TI\big) \,B\Big) \enspace,
\end{equation}
where we enforce the underlying physical model of temporal integration. 
This loss and all the following ones are self-supervised and do not require ground truth.

\boldparagraph{Time-consistency loss $\La_\Lt$} 
captures temporal smoothness of the sub-frame renderings, according to the prior knowledge of the problem.
We expect that renderings $\Ri$ will be similar for nearby $t$. Therefore, the similarity between renderings at two different points in time is defined as the maximum value of normalized cross-correlation over the image domain, which can be efficiently implemented on GPU using convolutions. To account for possible object translation, we apply zero-padding of 10\% of the image size to one of the renderings. The loss is defined as
\begin{equation}
\mathfont
	\label{eq:lt}
	\La_\Lt(\MySet{\Ri}) = 1 - \int_{0}^{1} \text{maxncc}(\Ri, \Ren_{\TI+\text{d}\TI}) \:\text{d}\TI \enspace.
\end{equation}


\boldparagraph{Sharpness loss $\La_\Ls$} 
enforces sharp image reconstruction, which is the main task of deblurring.
In the FMO setting, the object sharpness is assessed by its mask. Enforcing the mask to be binary is not ideal since we have mixed pixels at object boundaries. However, almost all pixels are expected to be close to zero or one. One mathematical way to express this statement is to minimize the per-pixel binary entropy $H_2$ averaged over the image domain $D$,
\begin{equation}
\mathfont
	\label{eq:ls}
	\begin{split}
	\La_\Ls(\MySet{\Mi}) = \int_{0}^{1} \frac{1}{|D|} \sum_{p \in D} H_2 \big(\Mi(p)\big) \:\text{d}\TI \enspace.
	\end{split}
\end{equation}

\boldparagraph{Latent learning loss $\La_\Ll$} 
models the latent space such that blurred images of the same FMO moving along the same trajectory but in front of different backgrounds generate the same latent representation.
We achieve this by training in such pairs of images, and compute the loss as
\begin{equation}
\mathfont
	\label{eq:ll}
	\La_\Ll(X^1, X^2) = \frac{1}{K} \| X^1 - X^2 \|_1 \enspace,
\end{equation}
where $X^1$ and $X^2$ are latent spaces generated by the first and the second images. All other losses, renderings and computations are done only on the first image,~\ie relating to $X^1$.
The latent learning loss especially helps in early training stages and stabilizes the training.

\boldparagraph{Joint loss}
Even perfectly optimized joint loss will not necessarily produce the ground truth renderings because some parts of the loss function are biased. 
For example, the time-consistency and sharpness losses are not minimized by the ground truth renderings since the ground truth masks are not binary on boundaries, and the appearance is not static. Whether the ground truth renderings are the global minimizers of the joint loss is an open research question.

\newcommand{\relme}{$\approx$}

Some parts of the joint loss 
can be loosely related to the energy minimization terms in the TbD~\cite{tbd} and TbD-3D~\cite{tbd3d} methods,~\eg appearance reconstruction loss~\relme~template-matching term~\cite{tbd}, image reconstruction loss~\relme~data likelihood~\cite{tbd,tbd3d}, time-consistency loss~\relme~regularization term enforcing similarity of the object in neighboring time intervals~\cite{tbd3d}. Other regularizers,~\eg total variation, are expected to be data-driven and learned by the network from the synthetic training set.
In comparison to our work, the previous methods are not neural network models, are handcrafted, computationally slow, and do not handle complex objects well (see Sec.~\ref{sec:rel-work}).

\section{Training and evaluation datasets}
\label{sec:train+eval}

\boldparagraph{Real-world datasets for evaluation} Only a small number of real-world annotated datasets with fast moving object exists.
The FMO dataset~\cite{fmo} contains 16 sports sequences, where FMOs are manually annotated with polygons. 
It lacks ground truth (GT) sharp appearances, segmentation masks, or even trajectories.
The TbD dataset~\cite{tbd} made a step further and captured 12 high-speed videos at 240~fps in raw format with full exposure. 
Subsequently, low-speed videos at 30~fps were created by temporal averaging.
Ground truth was generated from the high-speed video semi-manually by annotating the object in the first frame, applying the state-of-the-art tracker~\cite{Lukezic_2017_CVPR}, and correcting the mistakes.
Thus, ground truth for sharp appearances, masks, and full trajectory is provided in this dataset. 
However, the dataset contains only sports videos with mostly spherical objects and almost no appearance changes over time.
To address some of these shortcomings, the TbD-3D dataset~\cite{tbd3d} was recorded in the same fashion as TbD, but capturing objects that significantly change their appearance within one low-speed video frame. 
The dataset has only 10 sequences, and the objects are all but one spherical.
The recent Falling Objects dataset~\cite{kotera2020} is the first to contain objects with non-trivial shapes,~\eg box, pen, marker, cell, key, eraser. 
High-speed videos are provided, but no GT trajectories. 

We augmented the ground truth of~\cite{kotera2020} with trajectories by running the tracker~\cite{Lukezic_2017_CVPR} with annotations~--~we provide a manual box in the first frame, track the object, and correct the mistakes by re-initializing the tracker. We use this augmented dataset for testing and in the ablation study.

\renewcommand{\addimg}[1]{\includegraphics[width=0.09\textwidth]{#1}}

\newcommand{\AlgNameS}[1]{\raisebox{0.3em}{\rotatebox[origin=lt]{90}{\scriptsize #1}}}
\newcommand{\AlgNameL}[1]{\raisebox{1.5em}{\rotatebox[origin=lt]{0}{\small #1}}}

\newcommand{\makeRowSyn}[1]{ \raisebox{2.0em}{\multirow{3}{*}{\begin{tabular}{c}\addimg{#1_bgr} \\[0.5em]  \addimg{#1_im} \\ \end{tabular}}} & \AlgNameS{TbD-3D-O}  & \addimg{#1_tbd3dofm0} & \addimg{#1_tbd3dofm3}&  \addimg{#1_tbd3dofm5}&  \addimg{#1_tbd3dofm7}  \\
 & \AlgNameS{Ours} & \addimg{#1_estfm0} & \addimg{#1_estfm3} & \addimg{#1_estfm5} & \addimg{#1_estfm7}    \\
 & \AlgNameS{GT} & \addimg{#1_hs0} & \addimg{#1_hs3}& \addimg{#1_hs5} & \addimg{#1_hs7}   \\
\addlinespace[0.2em]
}

\begin{figure}
\centering
\footnotesize
\setlength{\tabcolsep}{0.03em} 
\renewcommand{\arraystretch}{0.15} 

\begin{tabular}{@{}c@{\hskip 0.5em}|@{\hskip 0.5em}ccccc@{}}

\makeRowSyn{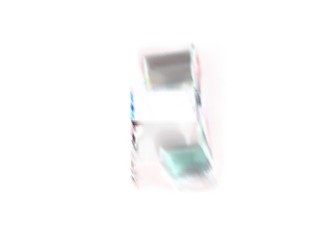} \\[-0.8em]
 Inputs ($B,I$) & & $F_{0}$ & $F_{0.25}$ & $F_{0.5}$ & $F_{0.75}$ \\
 
\end{tabular}
\vspace{-0.1em}
\caption{\textbf{FMO reconstruction on a validation image from the synthetic dataset.} We compare to the TbD-3D-Oracle~\cite{tbd3d} method (with ground truth trajectory). 
Flying table is synthetically superimposed on the image of a table tennis game.
}
\label{fig:results_syn}
\end{figure}

\boldparagraph{Synthetic dataset for training} 
is created due to the lack of a large and diverse real-world annotated dataset with FMOs.
To create a synthetic image, we use triplets: an \textit{object}, a 6D \textit{trajectory}, and a \textit{background} frame.
Then, we render the object along the given 6D trajectory using Blender Cycles~\cite{blender} to get a set of sub-frame renderings.
Finally, we apply the generalized formation model~\eqref{eq:blatting_int} to generate the low-speed frame showing the blurred FMO.

\begin{table}[t]
\centering
\small
\begin{center}
\setlength{\tabcolsep}{0.25em}
\newcommand{\CNameF}[1]{ \multirow{ 3}{*}{\rotatebox[origin=c]{90}{\scriptsize #1 }} }
\newcommand{\CNameA}[1]{ {\scriptsize #1 } }
\newcommand{\bmet}[1]{ \textbf{#1} }

\newcommand{\ApplyColor}[1]{\edef\x{\noexpand\cellcolor{lime!\PercentColor}}\x\textcolor{black}{#1}}
\newcommand{\ApplyGradient}[1]{%
    \IfDecimal{#1}{%
    \ifdim #1 pt > 0.75 pt
    \pgfmathsetmacro{\PercentColor}{100.0*max(min((#1-0.6)/(0.8-0.6),1.0),0.2)}\ApplyColor{\textbf{#1}}
    \else
    \pgfmathsetmacro{\PercentColor}{100.0*max(min((#1-0.6)/(0.8-0.6),1.0),0.2)}\ApplyColor{#1}
    \fi}
    {\CNameA{#1}}
}
\newcolumntype{R}{>{\collectcell\ApplyGradient}{c}<{\endcollectcell}}

\newcommand{\ApplyGradientP}[1]{%
    \IfDecimal{#1}{%
    \ifdim #1 pt > 26.3 pt
    \pgfmathsetmacro{\PercentColor}{100.0*max(min((#1-24)/(27-24),1.0),0.2)}\ApplyColor{\textbf{#1}}
    \else
    \pgfmathsetmacro{\PercentColor}{100.0*max(min((#1-24)/(27-24),1.0),0.2)}\ApplyColor{#1}
    \fi}
    {\CNameA{#1}}
}
\newcolumntype{P}{>{\collectcell\ApplyGradientP}{c}<{\endcollectcell}}

\newcommand{\ApplyGradientT}[1]{%
    \IfDecimal{#1}{%
    \ifdim #1 pt > 22.5 pt
    \pgfmathsetmacro{\PercentColor}{100.0*max(min((#1-21)/(22.5-21),1.0),0.2)}\ApplyColor{\textbf{#1}}
    \else
    \pgfmathsetmacro{\PercentColor}{100.0*max(min((#1-21)/(22.5-21),1.0),0.2)}\ApplyColor{#1}
    \fi}
    {\CNameA{#1}}
}
\newcolumntype{T}{>{\collectcell\ApplyGradientT}{c}<{\endcollectcell}}

\newcommand{\ApplyGradientI}[1]{%
    \IfDecimal{#1}{%
    \ifdim #1 pt > 0.7 pt
    \pgfmathsetmacro{\PercentColor}{100.0*max(min((#1-0.65)/(0.7-0.65),1.0),0.2)}\ApplyColor{\textbf{#1}}
    \else
    \pgfmathsetmacro{\PercentColor}{100.0*max(min((#1-0.65)/(0.7-0.65),1.0),0.2)}\ApplyColor{#1}
    \fi}
    {\CNameA{#1}}
}
\newcolumntype{I}{>{\collectcell\ApplyGradientI}{c}<{\endcollectcell}}
\newcommand{\yes}{\textcolor{black}{\checkmark}}
\newcommand{\noo}{\textcolor{red}{\xmark}}
\newcommand{\lss}[1]{\footnotesize $\La_#1$}

\begin{tabular}{cccccTTPRI}
\specialrule{.1em}{.05em}{.05em} 
\CNameF{Appear. rec.} & \CNameF{Image rec.} & \CNameF{Time-cons.} & \CNameF{Latent learn.} & \CNameF{Sharpness} &  \multicolumn{5}{c}{} \\
 & & & & & & & & & \\

 & & & & & \multicolumn{1}{c}{\small Train} & \multicolumn{1}{c}{\small Val} & \multicolumn{3}{c}{\small Test~--~Falling Obj.~\cite{kotera2020}} \\
\cmidrule(lr){6-6}
\cmidrule(lr){7-7}
\cmidrule(lr){8-10}


 \lss{\Lfm} & \lss{\Li} & \lss{\Lt} & \lss{\Ll} & \lss{\Ls} & PSNR$\uparrow$ &  PSNR$\uparrow$ &  PSNR$\uparrow$ &  SSIM$\uparrow$ & TIoU$\uparrow$ \\ 
 \hline

\multicolumn{5}{l}{\footnotesize TbD-3D~\cite{tbd3d}} & 6.58 & 6.59 & 23.0 & .695 & .545 \\ 
\yes & \noo & \noo & \noo & \noo & 23.0 & 22.6 & 24.8 & .691 & .684 \\
\yes & \yes & \noo & \noo & \noo & 22.5 & 22.2 & 25.5 & .705 & .653 \\
\noo & \yes & \yes & \yes & \yes & 11.6 & 10.9 & 19.7 & .459 & .347 \\ 
\yes & \noo & \yes & \yes & \yes & 12.3 & 12.2 & 17.5 & .362 & .489 \\
\yes & \yes & \noo & \yes & \yes & 21.6 & 21.5 & 25.8 & .743 & .673 \\
\yes & \yes & \yes & \noo & \yes & 22.2 & 22.0 & 26.1 & .739 & .678 \\
\yes & \yes & \yes & \yes & \noo & 22.4 & 22.2 & 26.4 & .750 & .676 \\
\yes & \yes & \yes & \yes & \yes & 22.5 & 22.4 & 26.4 & .753 & .703 \\
                
\specialrule{.1em}{.05em}{.05em} 

\end{tabular}


\end{center}
\vspace{-0.5em}
\caption{\textbf{Ablation study of DeFMO.} 
Training and validation datasets are generated synthetically. 
Training without additional self-supervised losses overfits to the training set (rows 2, 3). 
Values are color-coded, darker is better.
}
\label{tbl:ablation}
\end{table}

\textit{Objects} are sampled from 3D models of the 50 largest classes of the ShapeNet~\cite{shapenet2015} dataset, each class is represented uniformly. 
Since most ShapeNet objects are not well-textured,
we apply DTD~\cite{dtd} textures,
as in the dataset creation for the Neural Voxel Renderer~\cite{Rematas2020}.
The textures are split into 1600 for training and 200 for validation. 

\textit{Trajectories} 
are sampled uniformly as linear with the displacement in the range between 0.5 and 2 object sizes in $x,y$ image directions, and between 0 and 0.2 sizes in $z$ direction towards the camera.
3D rotations are sampled with a maximal rotation change of $30^{\circ}$ in each direction.
For rendering, the trajectory is discretized into $N=24$ equal parts.

\textit{Backgrounds} are sampled from the VOT~\cite{VOT_TPAMI} sequences for training, and from Sports-1M~\cite{KarpathyCVPR14} for validation. 
For the latent space learning, the foreground image is synthesized on a pair of backgrounds.
The background $B$ that we use as input to the method is estimated as a median over 5 previous frames. 
This way, the method gets exposed to complex non-static backgrounds as the VOT dataset has a variety of dynamic scenes but no other FMOs~\cite{fmo}.

In total, we generate 50,000 training and 1,000 validation images (example in Fig.~\ref{fig:results_syn}). 
The synthetic dataset is more challenging than a real-world dataset (ablation study Table~\ref{tbl:ablation}). 
The reported metrics are better on the test dataset (Falling Objects~\cite{kotera2018}) than on the synthetic training dataset, which has more complex objects and textures. 

\setcounter{footnote}{0}
\newcommand{\addtypimg}[1]{ \raisebox{-0.15em}{\multirow{3}{*}{\includegraphics[width=0.11\linewidth]{imgs/typical/#1}} }}

\begin{table*}[t]
\centering
\small
\newcommand{\CNameFA}[1]{ \raisebox{0.1em}{\multirow{ 3}{*}{\rotatebox[origin=l]{90}{\scriptsize #1 }}} }
\newcommand{\CName}[1]{ {\footnotesize #1 } }
\setlength{\tabcolsep}{2.2mm} 
\renewcommand{\arraystretch}{1.2} 
\begin{center}
\begin{tabular}{@{}c@{\hskip 0.2em}c@{}rcccccccc@{}}

\specialrule{.1em}{.05em}{.05em} 

 \multirow{2}{*}[-5pt]{\rotatebox[origin=l]{90}{\footnotesize Dataset}} & \multirow{2}{*}{\CName{Typical Object}} & \multirow{2}{*}{\CName{Score}} & \multicolumn{2}{c}{ \CName{Inputs} } & \multicolumn{4}{c}{  \CName{Compared Methods} }  &  \CName{Proposed} &  \CName{Traj. Oracle} \\
  \cmidrule(lr){4-5} \cmidrule(lr){6-9} \cmidrule(lr){10-10} \cmidrule(lr){11-11}

 &  &  & \CName{$B$} & \CName{$I$} & \CName{Jin~\etal~\cite{Jin_2018_CVPR}} & \CName{Debl.GAN~\cite{Kupyn_2019_ICCV}} & \CName{TbD~\cite{tbd}} & \CName{TbD-3D~\cite{tbd3d}} & \CName{DeFMO} & \CName{TbD-3D-Or.}  \\ 
 
 \hline

 \CNameFA{Falling~\cite{kotera2020}} & \addtypimg{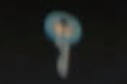} & TIoU$\uparrow$ & N/A & N/A & N/A & N/A & 0.539 & 0.539 & \winner{0.684}  & 1.000 \\
 
 & & PSNR$\uparrow$ & 19.71 & 23.76 & 23.54 & 23.36 & 20.53 & 23.42  & \winner{26.83} & 23.38 \\
  
 & & SSIM$\uparrow$ & 0.456 & 0.594 & 0.575 & 0.588 & 0.591 & 0.671 & \winner{0.753} & 0.692  \\

\hline

\CNameFA{TbD-3D~\cite{tbd3d}} & \addtypimg{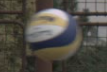} &  TIoU$\uparrow$ & N/A & N/A & N/A & N/A  & 0.598 & 0.598  & \winner{0.879} & 1.000 \\

 & & PSNR$\uparrow$ & 19.81 & 24.80 & 24.52 & 23.58  & 18.84 & 23.13 & \winner{26.23} & 24.84  \\
  
 & & SSIM$\uparrow$ & 0.426 & 0.640 & 0.590 & 0.603 & 0.504 & 0.651 &  \winner{0.699} & \oracle{0.705}  \\

\hline

\CNameFA{TbD~\cite{tbd}} & \addtypimg{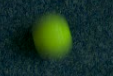} &  TIoU$\uparrow$ & N/A & N/A & N/A & N/A  & 0.542 & 0.542  & \winner{0.550} & 1.000 \\

 & & PSNR$\uparrow$ & 21.48 & 25.06 & 24.90  & 24.27  & 23.22 & 25.21  & \winner{25.57} &  \oracle{26.36}  \\

 & & SSIM$\uparrow$ & 0.466 & 0.568 & 0.530 & 0.537 & 0.605  & \winner{0.674} &  0.602 &  \oracle{0.712} \\
 \hline 

\multicolumn{3}{c}{Runtime (on $240 \times 320$)} & N/A & N/A & 2 fps & 10 fps & 0.01 fps & 0.001 fps  & \winner{20 fps} & 0.001 fps \\

\specialrule{.1em}{.05em}{.05em} 

\end{tabular}
\end{center}
\vspace{-0.9em}
\caption[Something]{\textbf{Evaluation on:} Falling Objects~\cite{kotera2020}, TbD-3D~\cite{tbd3d}, TbD~\cite{tbd} datasets\footnotemark. 
The datasets are sorted by decreasing difficulty: arbitrary shaped and textured~\cite{kotera2020}, mostly spherical but significantly textured~\cite{tbd3d}, and mostly spherical and uniformly colored objects~\cite{tbd}.
DeFMO is superior to the compared methods by a wide margin on all datasets except for the easiest~\cite{tbd} since the TbD(-3D) methods~\cite{tbd,tbd3d} are specifically designed for such easy cases. TbD-3D-Oracle~\cite{tbd3d} is given ground truth trajectories and estimates only sub-frame appearance. }
\label{tbl:datasets}
\end{table*}



\boldparagraph{Training settings} 
Both the encoder and the rendering networks are based on ResNet~\cite{resnet} with batch norm~\cite{batchnorm} and ReLU~\cite{relu}. 
The encoder is ResNet-50 cropped after the fourth down-sampling operator, pre-trained on ImageNet~\cite{ILSVRC15}. 
The latent space is up-scaled by $2$ using pixel shuffle~\cite{pixelshuffle} four times, each followed by ResNet bottleneck (\ie $1024,256,64,16,4$ channels). 
We used ADAM~\cite{adam} optimizer with fixed learning rate $10^{-3}$. 
For the input image resolution $w \times h$, the resolution of the latent space is $2048 \times w / 16 \times h / 16 = K$ (due to 4 down-sampling operators).
The input to the rendering network is augmented by copying the time index channel $t$ along the first dimension, leading to resolution $2049 \times w / 16 \times h / 16$.
In experiments, we set $w = 320$ and $h = 240$. 
Loss weights are set as $\Lp_\Li, \Lp_\Ls, \Lp_\Ll = 1$ and $\Lp_\Lt = 5$.
The encoder and the rendering networks have 23.5M and 20.1M parameters, respectively. 
The model is implemented in PyTorch~\cite{NEURIPS2019_9015}, trained in 24 mini-batches on 3 nVidia 16GB GPUs. 
We trained for 50 epochs, which took approximately four days. 
For evaluation, the method runs in real-time on a single GPU and works for various resolutions and sub-frame time indices.


\section{Evaluation}
\label{sec:eval}

\footnotetext{\href{https://github.com/rozumden/fmo-deblurring-benchmark/}{https://github.com/rozumden/fmo-deblurring-benchmark/}\label{note1}}
We evaluated DeFMO on several datasets\textsuperscript{\ref{note1}} (Sec.~\ref{sec:train+eval}) and compared it to the state of the art. 
Evaluation metrics are PSNR (peak signal-to-noise ratio), SSIM (structural similarity index measure), and TIoU (Intersection over Union averaged along the trajectory)~\cite{tbd}. 
We compare to FMO deblurring methods based on energy minimization: TbD~\cite{tbd}, TbD-3D~\cite{tbd3d}, and TbD-3D-Oracle (where we provide the ground truth trajectory). All FMO deblurring methods and ours use the same estimate of the background~--~a median of the previous 5 frames.
We also compare to two 
state-of-the-art 
generic deblurring and temporal super-resolution methods: DeblurGAN-v2~\cite{Kupyn_2019_ICCV} (generates a single output, compared to the best aligned high-speed frame) and Jin~\etal~\cite{Jin_2018_CVPR} (generates a mini-video, the same comparison to GT as for DeFMO).
We do not compare to the method~\cite{kotera2020} since it covers a very special case of a constant planar appearance with 2D rotation parallel to the image plane, and for the general case, it is inferior to TbD-3D~\cite{tbd3d}.
The authors of~\cite{kotera2020} introduced the Falling Objects dataset but evaluated only qualitatively on three chosen frames
where the object appearance was constant.

\newcommand{\addimgT}[1]{\includegraphics[trim={2cm 1cm 0 1cm},clip,height=0.29\linewidth]{#1}}
\newcommand{\addimgTF}[1]{\includegraphics[trim={0cm 3cm 0 1cm},clip,height=0.29\linewidth]{#1}}

\newcommand{\AlgNameT}[1]{\raisebox{2.0em}{\rotatebox[origin=lt]{90}{\scriptsize #1}}}

\begin{figure}
\vspace{-5pt}
\centering
\setlength{\tabcolsep}{0.25em} 
\renewcommand{\arraystretch}{0.25} 
\begin{tabular}{@{}ccc@{}}

\AlgNameT{Ground truth} & \addimgT{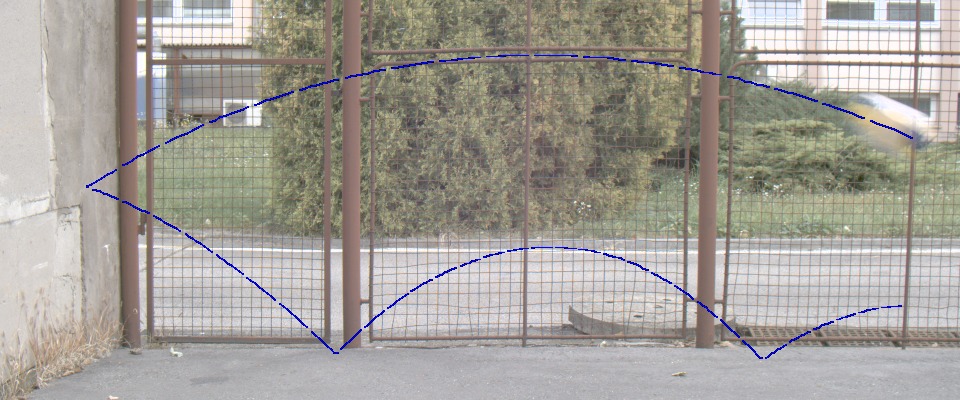} & \addimgTF{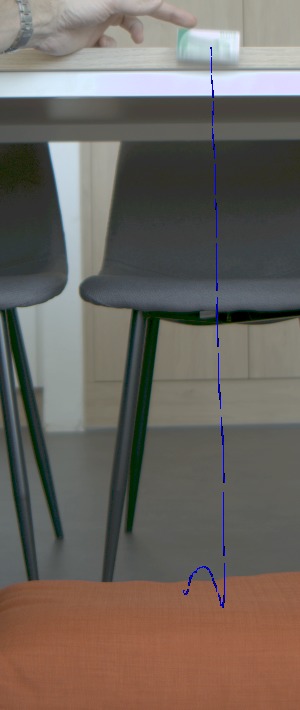} \\
\AlgNameT{DeFMO (Ours)} & \addimgT{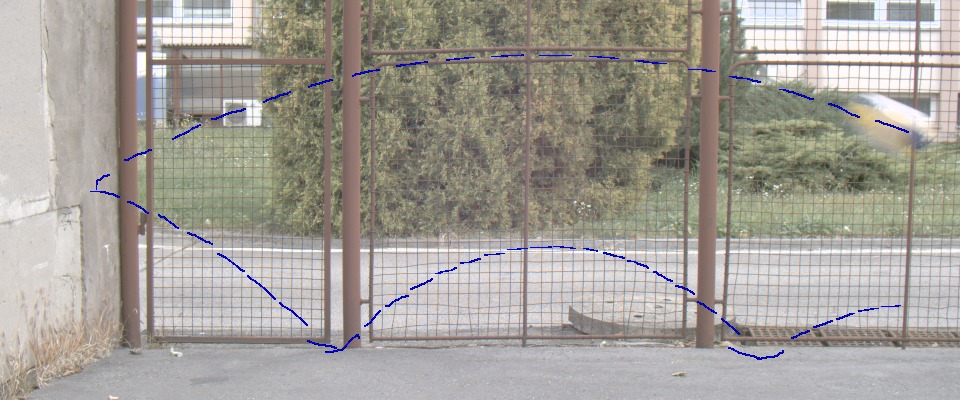} & \addimgTF{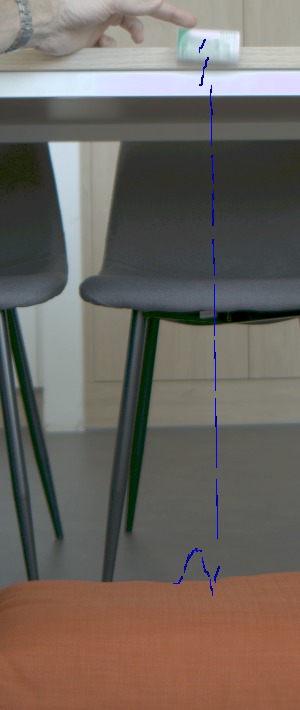} \\
\AlgNameT{TbD~\cite{tbd}} & \addimgT{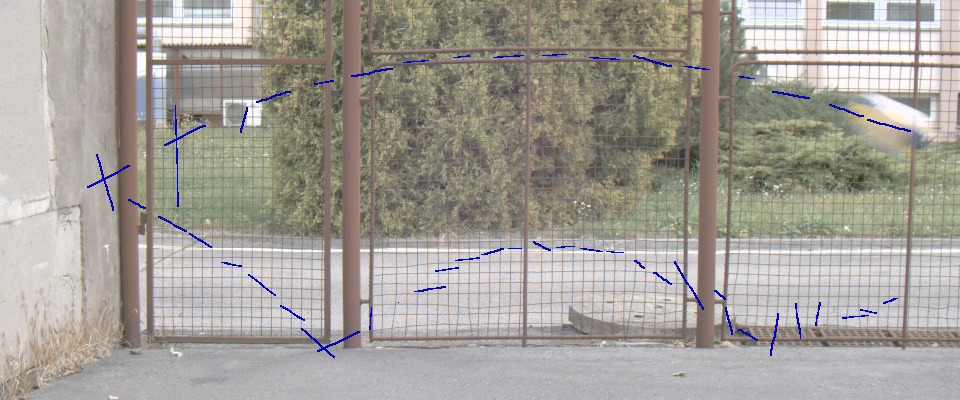}  & \addimgTF{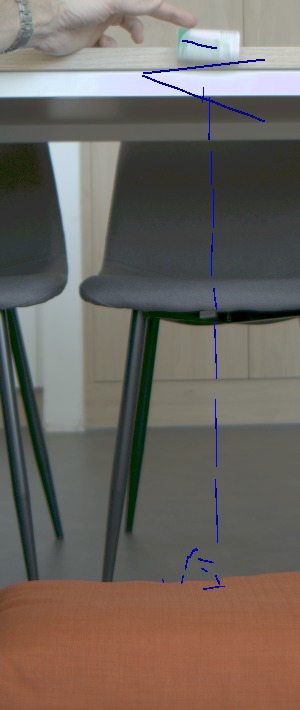} \\[0.3em]
\end{tabular}

\caption{\textbf{Trajectory estimation} on sequences from the TbD-3D dataset~\cite{tbd3d} (left) and Falling Objects dataset~\cite{kotera2020} (right). 
}
\label{fig:traj}
\vspace{-9pt}
\end{figure}

\renewcommand{\addimg}[1]{\includegraphics[width=0.057\linewidth]{#1.jpg}}

\newcommand{\AlgName}[1]{\rotatebox[origin=lt]{90}{\scriptsize #1}}

\newcommand{\makeRowFull}[2]{ \raisebox{0.0em}{\multirow{3}{*}{\rotatebox[origin=c]{90}{\scriptsize #2}}} & 
\raisebox{1.5em}{\multirow{3}{*}{ \begin{tabular}{c}\addimg{#1_bgr} \\[0.5em]  \addimg{#1_im} \\ \end{tabular}}} & 
\AlgName{Orac} & \addimg{#1_tbd3do0} & \addimg{#1_tbd3do1} & \addimg{#1_tbd3do2} & \addimg{#1_tbd3do3}& \addimg{#1_tbd3do4}& \addimg{#1_tbd3do5}& \addimg{#1_tbd3do6}& \addimg{#1_tbd3do7}  & \addimg{#1_tbd3dom0} & \addimg{#1_tbd3dom1} & \addimg{#1_tbd3dom2} & \addimg{#1_tbd3dom3}& \addimg{#1_tbd3dom4}& \addimg{#1_tbd3dom5}& \addimg{#1_tbd3dom6}& \addimg{#1_tbd3dom7} \\
&  & \AlgName{Ours} &  \addimg{#1_est0} & \addimg{#1_est1} & \addimg{#1_est2} & \addimg{#1_est3} & \addimg{#1_est4} & \addimg{#1_est5} & \addimg{#1_est6}& \addimg{#1_est7} & \addimg{#1_estm0} & \addimg{#1_estm1} & \addimg{#1_estm2} & \addimg{#1_estm3} & \addimg{#1_estm4} & \addimg{#1_estm5} & \addimg{#1_estm6}& \addimg{#1_estm7}  \\
 & & \AlgName{GT} & \addimg{#1_hs0} & \addimg{#1_hs1} & \addimg{#1_hs2} & \addimg{#1_hs3} & \addimg{#1_hs4} & \addimg{#1_hs5} & \addimg{#1_hs6}& \addimg{#1_hs7} & \addimg{#1_hsm0} & \addimg{#1_hsm1} & \addimg{#1_hsm2} & \addimg{#1_hsm3} & \addimg{#1_hsm4} & \addimg{#1_hsm5} & \addimg{#1_hsm6}& \addimg{#1_hsm7} \\
\addlinespace[0.01em]
}

\begin{figure*}
\centering
\footnotesize
\setlength{\tabcolsep}{0.01em} 
\renewcommand{\arraystretch}{0.1} 

\begin{tabular}{@{}cc@{\hskip 0.2em}ccccccccccccccccc@{}}

\makeRowFull{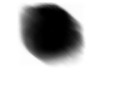}{volleyball}
\makeRowFull{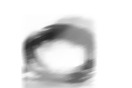}{aerobie}
\makeRowFull{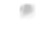}{shuttlecock}
\makeRowFull{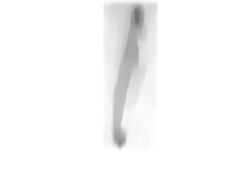}{pen}

\addlinespace[0.3em]
\multicolumn{3}{c}{Inputs ($B,I$)} & 
\multicolumn{8}{c}{Temporal super-resolution} &
\multicolumn{8}{c}{Masks $M_t$} \\[0.6em]

\end{tabular}
\caption{\textbf{Temporal super-resolution on selected sequences from test datasets.} 
We compare to the TbD-3D-Oracle~\cite{tbd3d} with the manually provided ground truth trajectory from a high-speed footage (GT). Ground truth masks are computed as a difference image between the GT sub-frame and the background. The proposed DeFMO method estimates everything just from inputs on the left.}
\label{fig:results_full}
\end{figure*}

The high-speed video is available at 8 times higher frame rate than the low-speed video, both at full exposure. Therefore, we generate full exposure ($\expos = 1$) temporal super-resolution for quantitative evaluation to match the high-speed 
frames. 
As discussed in Sec.~\ref{sec:model}, we generate $\{I_{k/8:(k+1)/8}\}_{k=0}^{7}$ according to~\eqref{eq:blatting_int}, and discretize each integral by $5$ parts. 
For qualitative results, we reconstruct objects as sharp as possible and visualize zero exposure temporal super-resolution.
Since our temporal super-resolution task is done from a single image, the direction of time is ambiguous. 
Hence, we compute scores for both directions and report the best one (for all methods).
FMOs are retrieved and approximately localized by the bounding box using the FMO detector~\cite{fmo}.
Sub-frame trajectory (Fig.~\ref{fig:traj}) is estimated as the center of mass of the generated masks $\Mi$. 

\boldparagraph{Falling Objects dataset} Most previous methods fail on this challenging dataset (Table~\ref{tbl:datasets}, top block~\cite{kotera2020}), mainly because the objects change their appearance too much within one frame (Fig.~\ref{fig:results}). DeFMO outperforms all others in all metrics on this dataset. It even outperforms TbD-3D-Oracle, although that method requires GT trajectories as input, which are not available in real-world scenarios.

\boldparagraph{TbD-3D dataset}
Table~\ref{tbl:datasets} (middle block~\cite{tbd3d}) shows that DeFMO outperforms all other methods not using the ground truth trajectory.
DeFMO compares well even against TbD-3D-Oracle (better PSNR, slightly worse SSIM). 

\boldparagraph{TbD dataset}
contains objects that are mostly spherical and with constant appearance.
In this simplistic setting, DeFMO is slightly worse than TbD(-3D) methods on the SSIM metric (Table~\ref{tbl:datasets}, bottom block~\cite{tbd}).
On PSNR and TIoU metrics, DeFMO is the best-performing method. 
Note that TbD(-3D) methods are specifically designed for such simple objects and work well there. 
Our method is more general and does not have assumptions of spherical or constant objects. 
However, DeFMO is several orders of magnitude faster but comparable in performance there. 
One way to improve the performance on such objects is to generate a synthetic training set with objects of constant appearance.

\newcommand{\addimgW}[1]{\includegraphics[height=0.116\linewidth]{#1.jpg}}
\newcommand{\addimgWII}[1]{\includegraphics[height=0.120\linewidth]{#1.jpg}}
\newcommand{\addimgWIII}[1]{\includegraphics[height=0.108\linewidth]{#1.jpg}}

\begin{figure}
\centering
\scriptsize
\setlength{\tabcolsep}{0.03em} 
\renewcommand{\arraystretch}{0.5} 
\newcommand{\BlockSkip}{\hskip 0.4em}
\begin{tabular}{@{}cc@{\BlockSkip}ccc@{\BlockSkip}c@{}}
\addimgWII{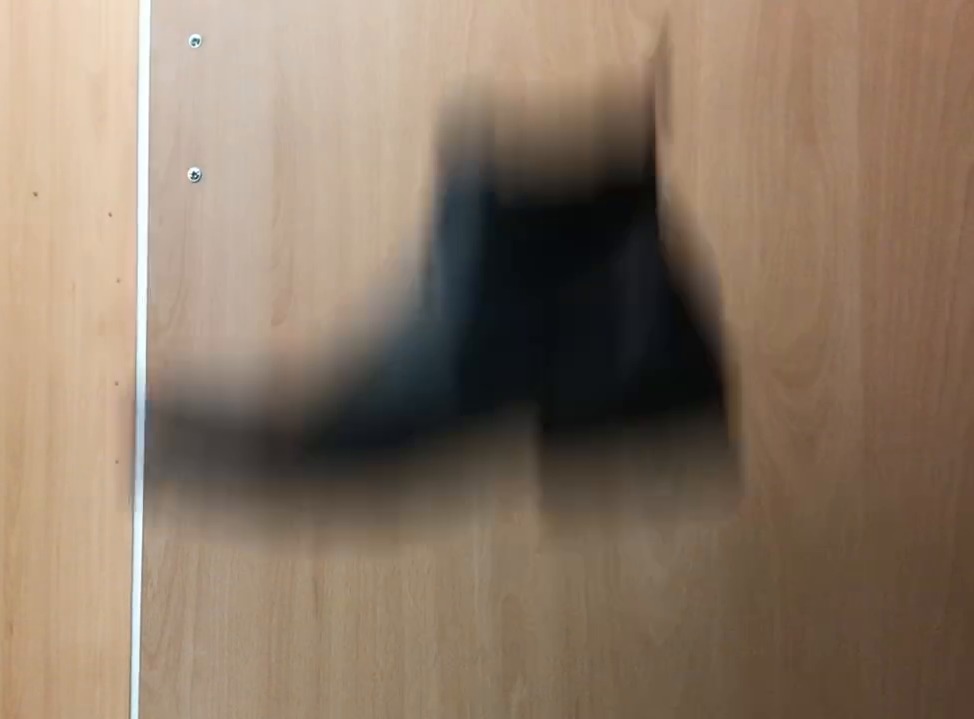} & \addimgWII{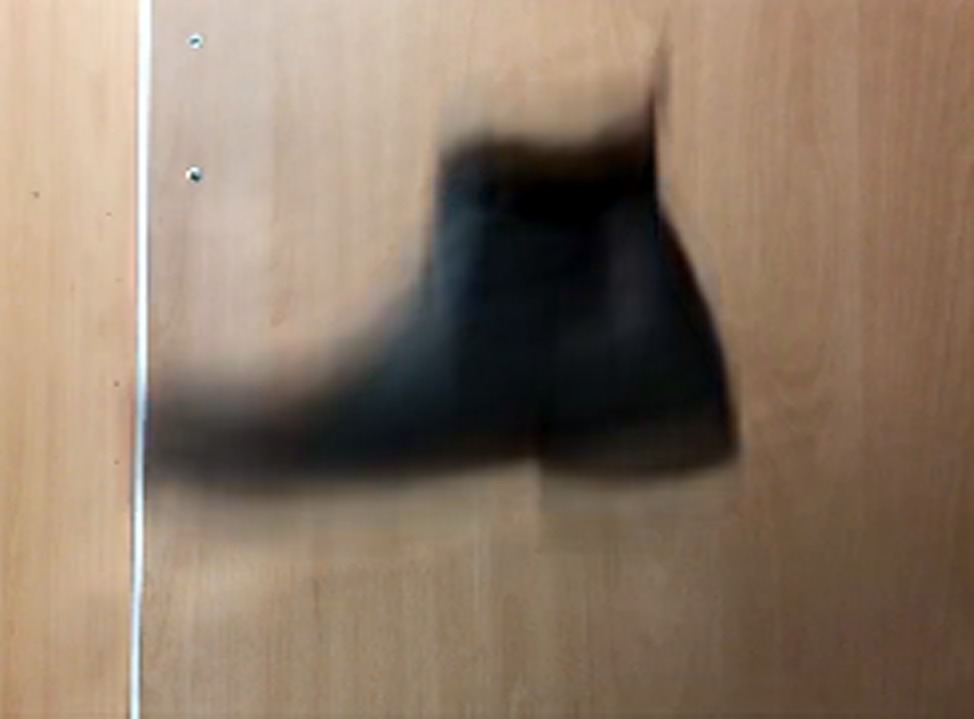} & \addimgWII{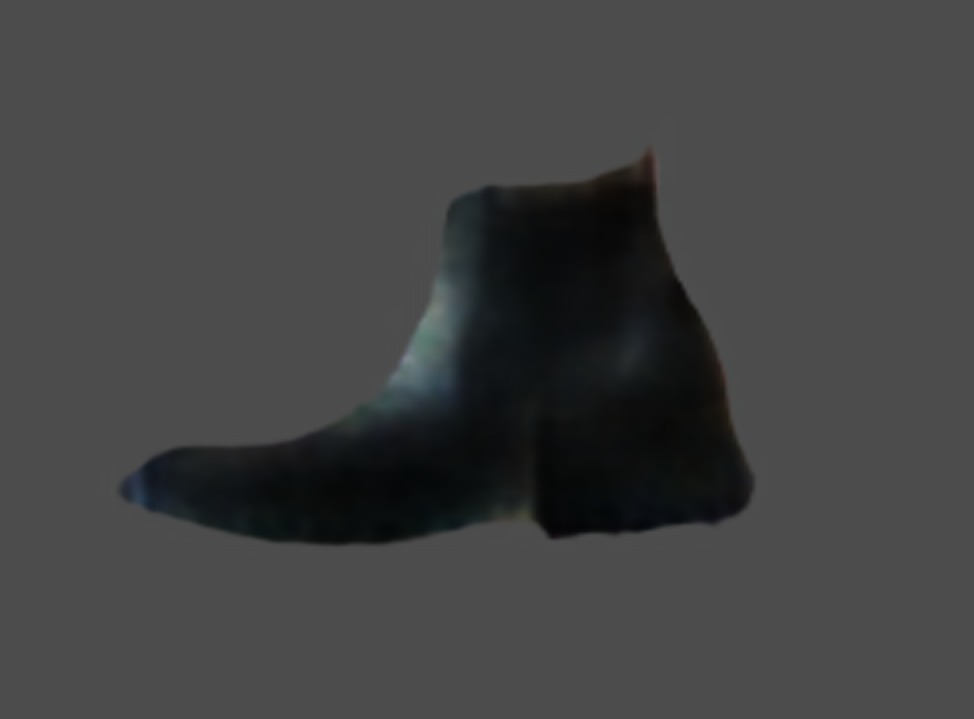}  & \addimgWII{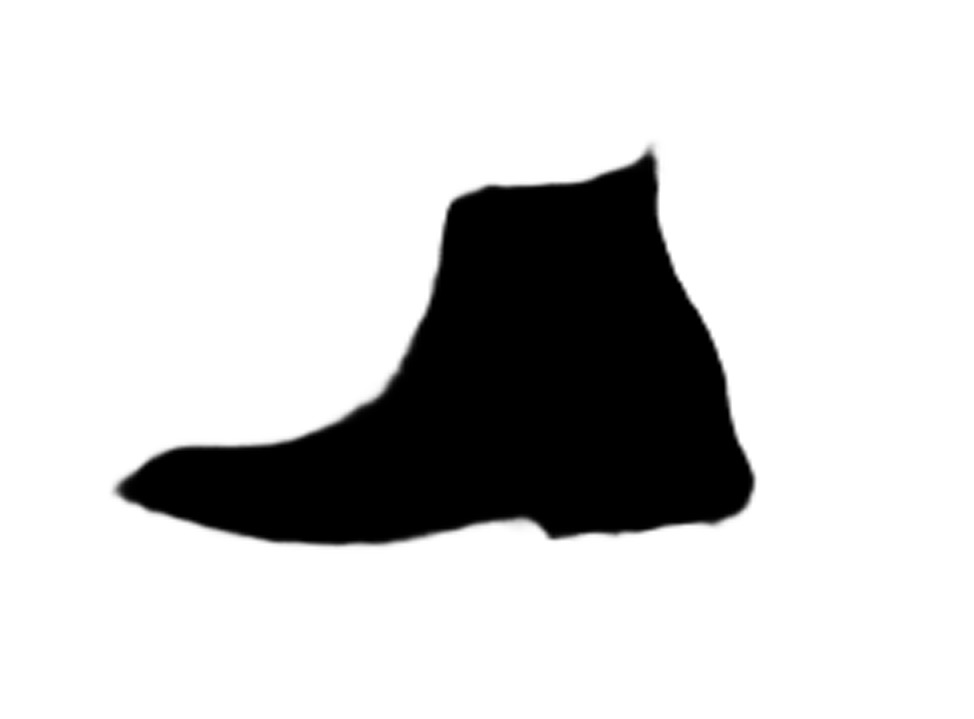} & \addimgWII{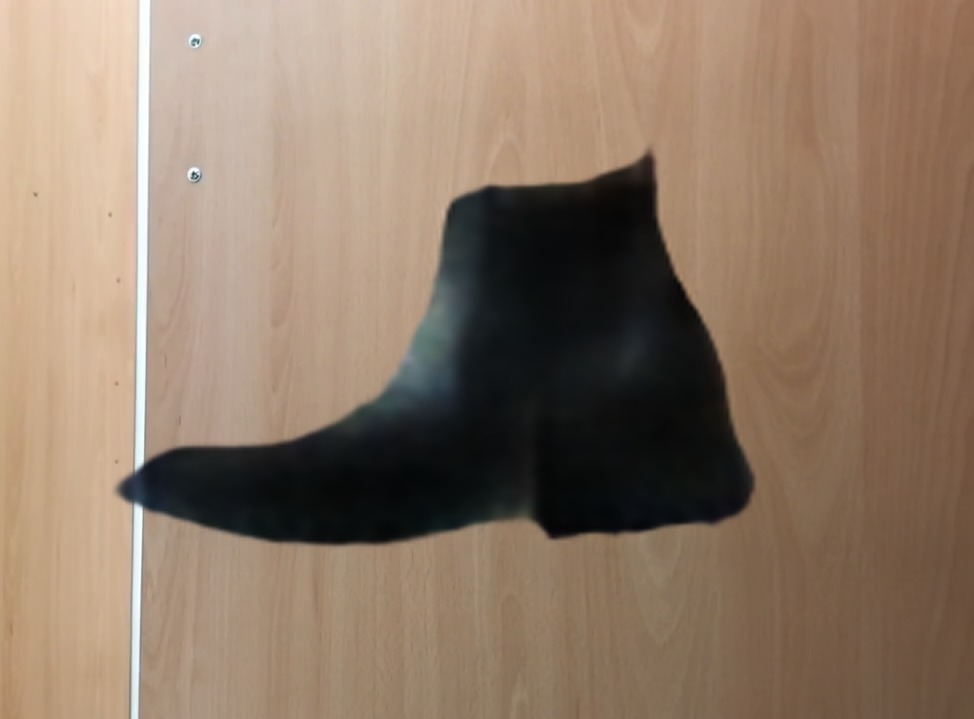} & 
\addimgWII{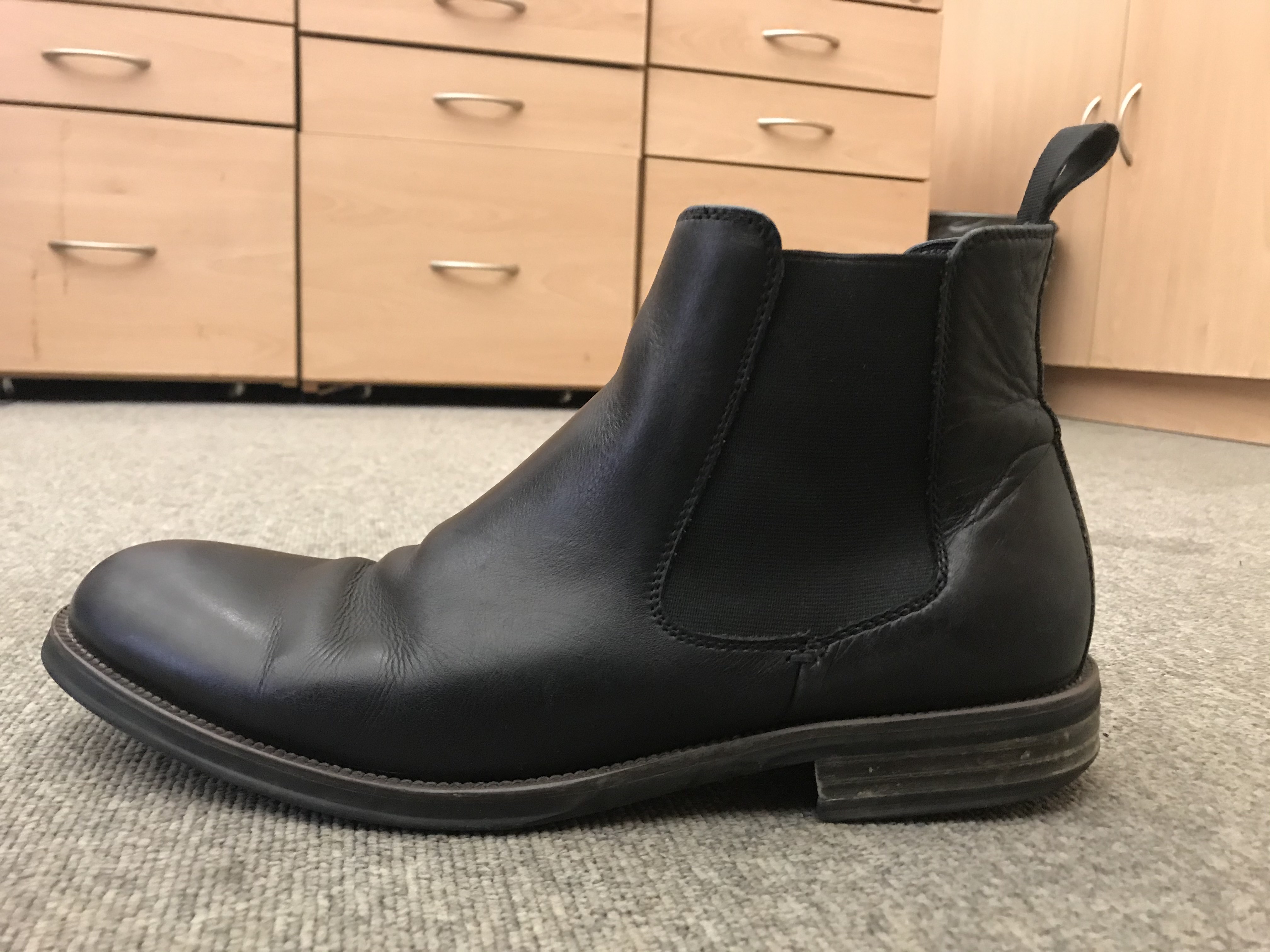} \\
\addimgW{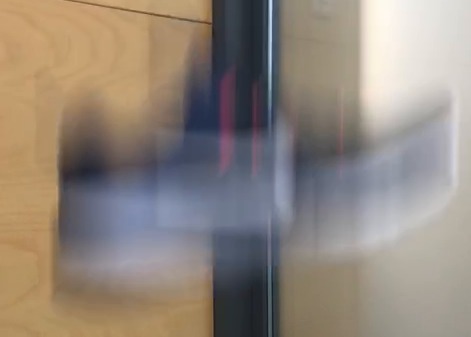} & \addimgW{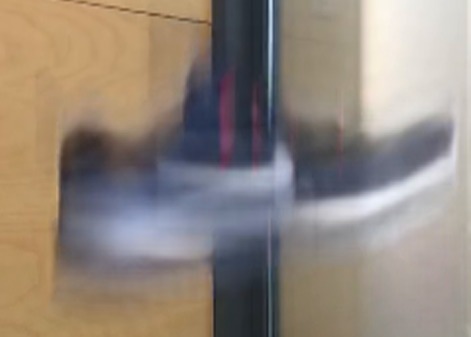} & \addimgW{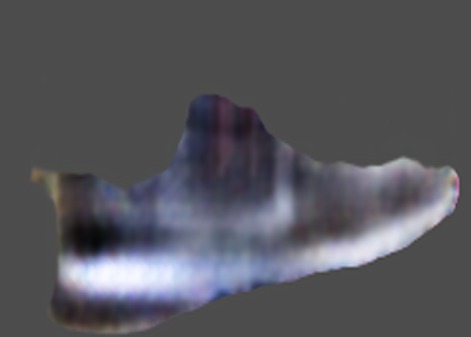}  & \addimgW{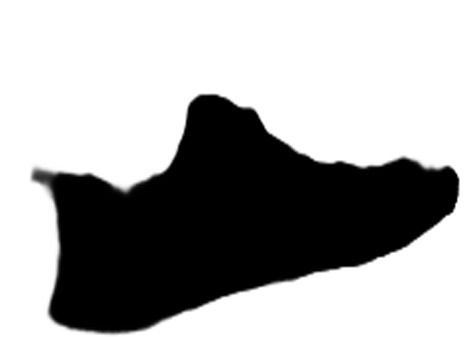} & \addimgW{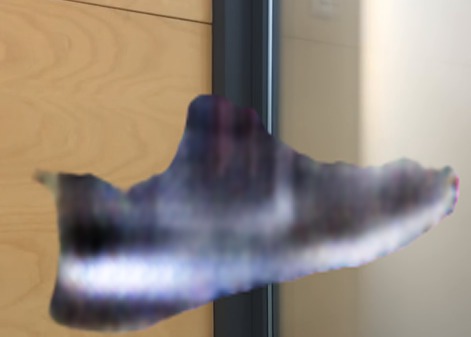} & 
\addimgW{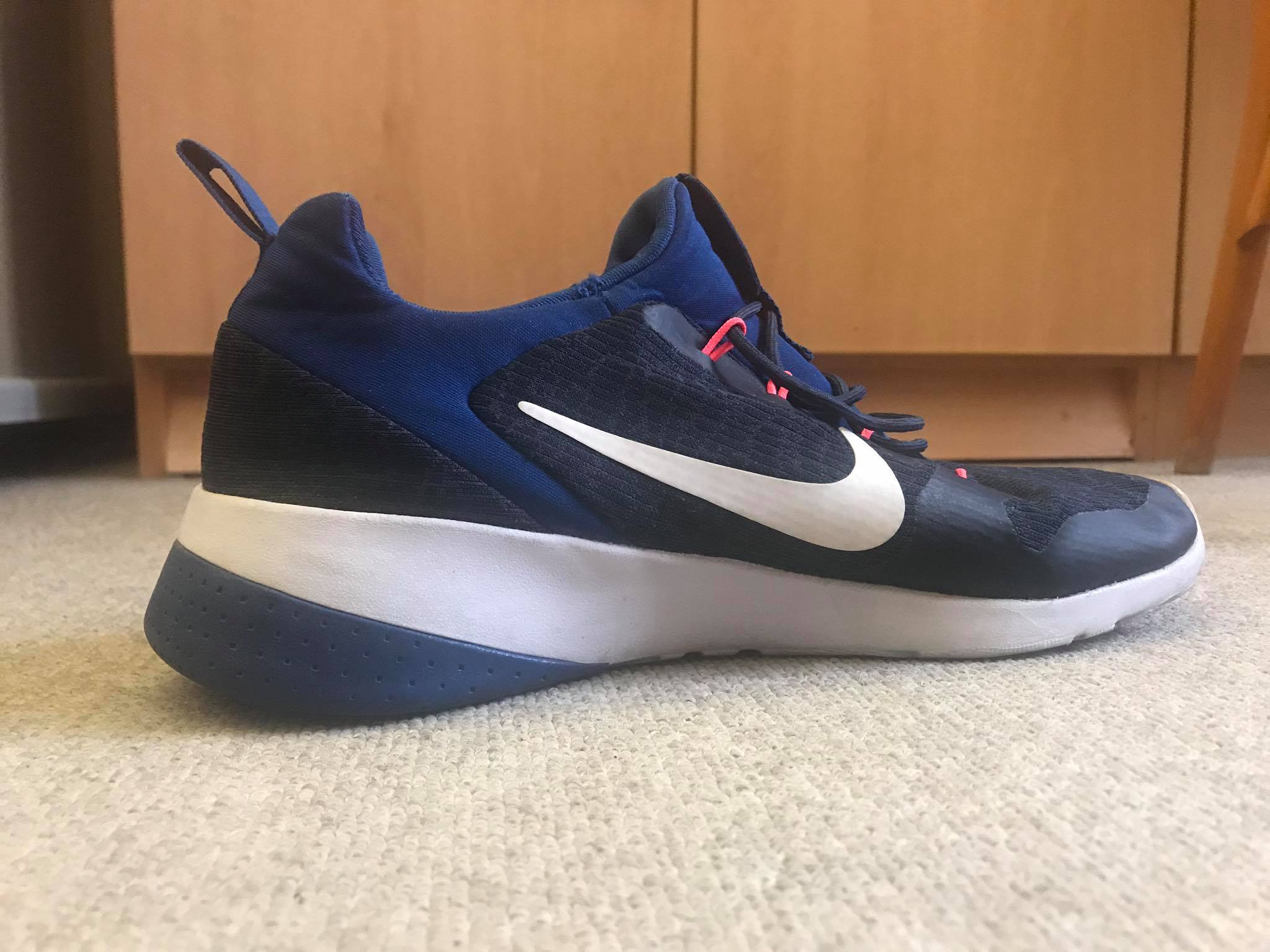} \\
\addimgWIII{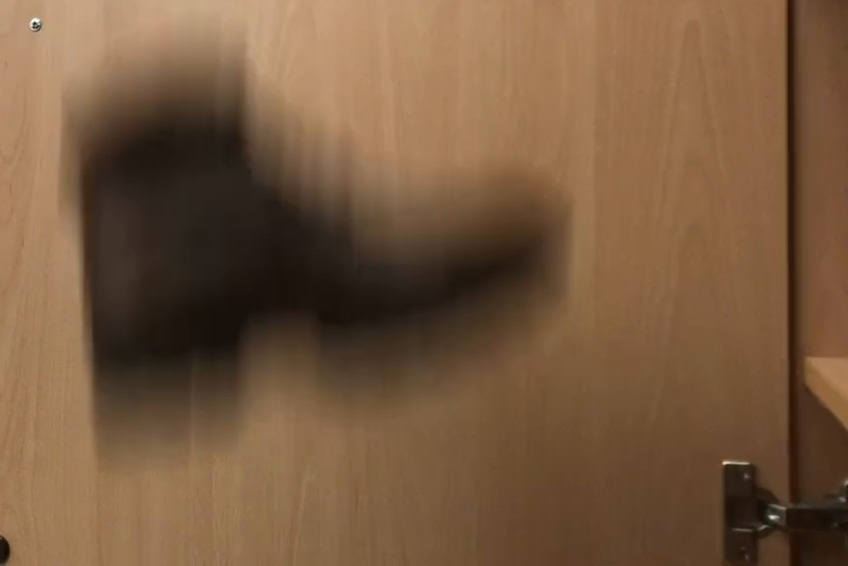} & \addimgWIII{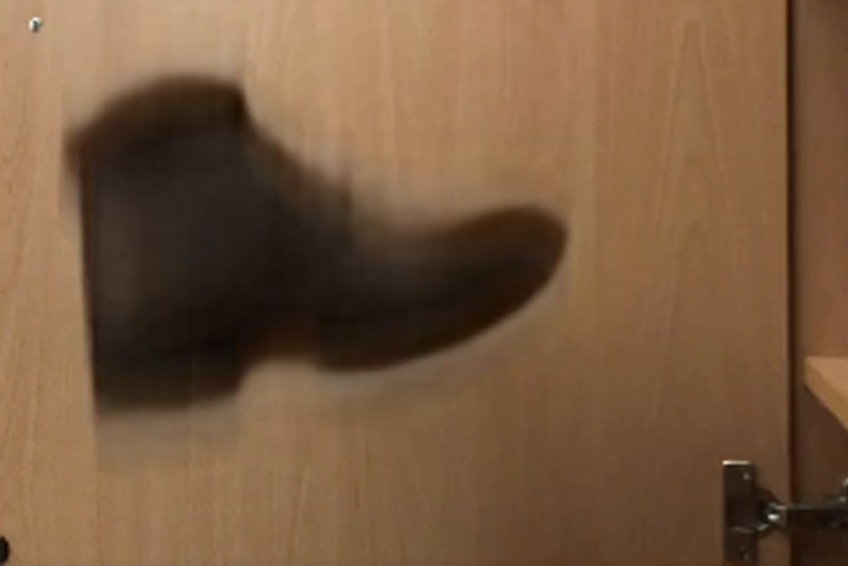} & \addimgWIII{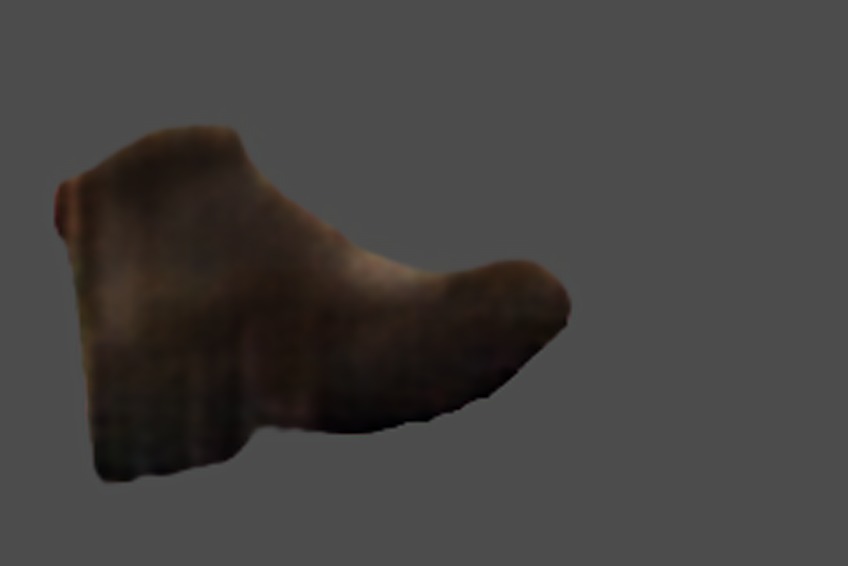}  & \addimgWIII{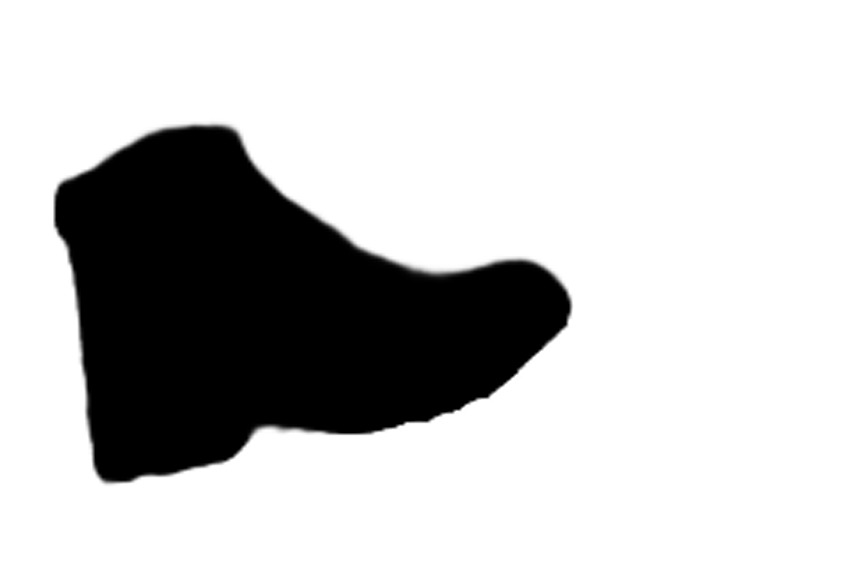} & \addimgWIII{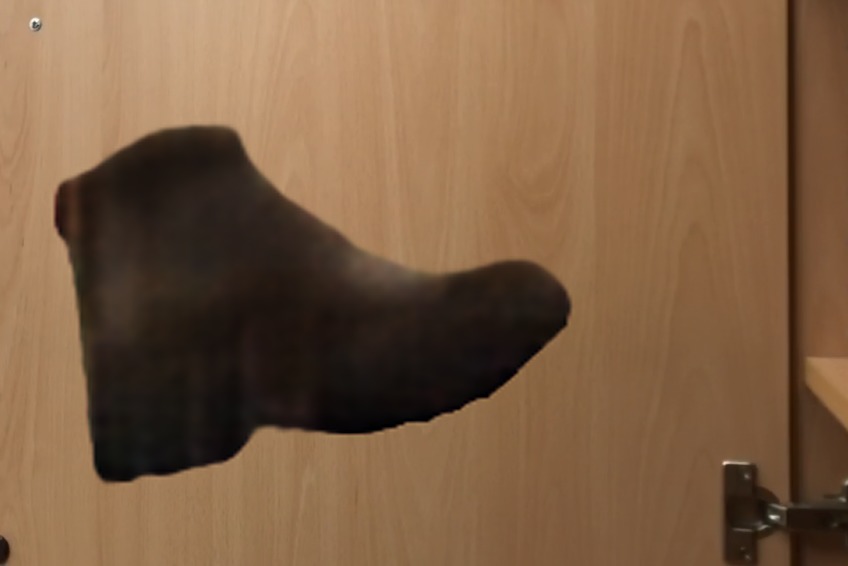} & 
\addimgWIII{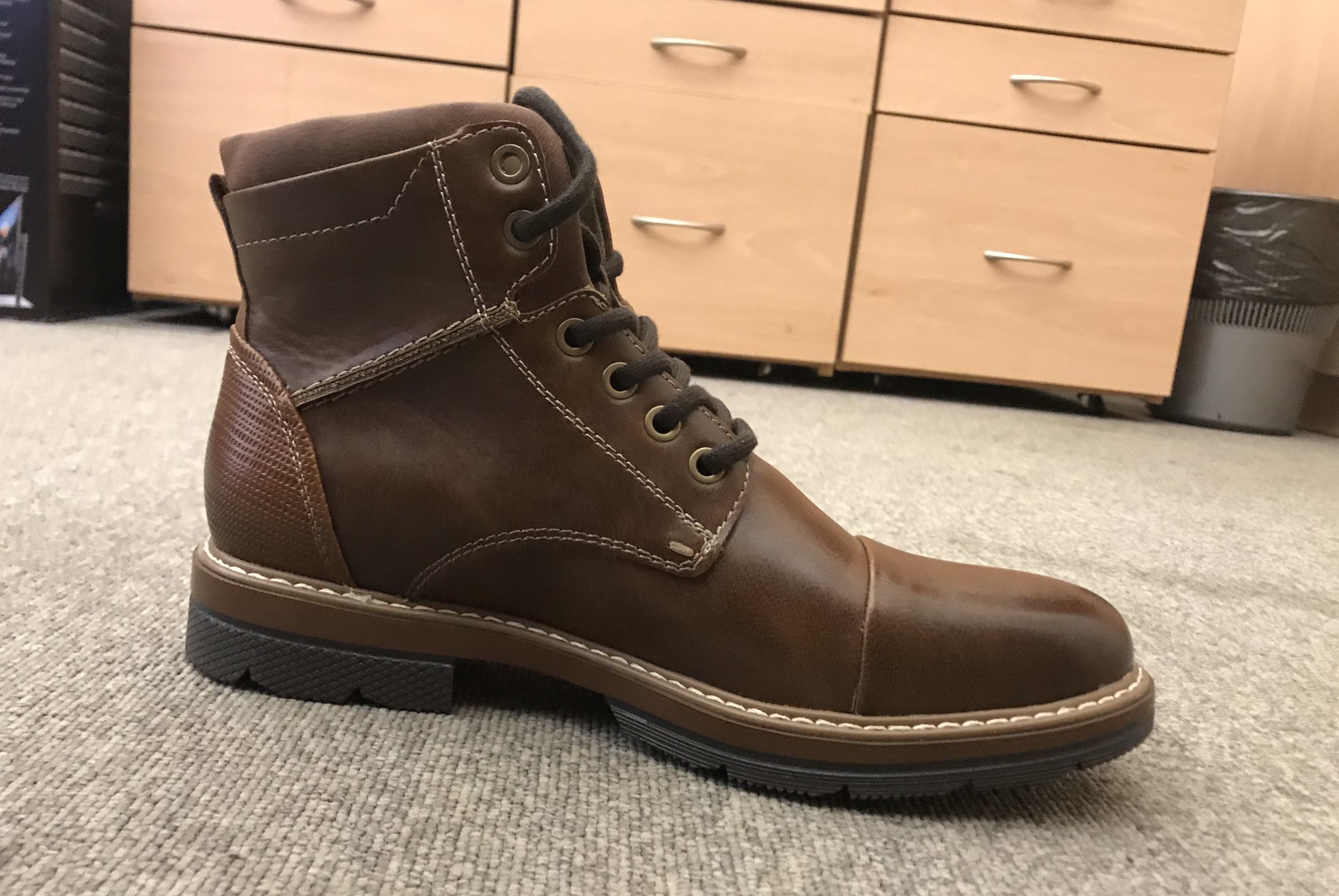} \\
Input $I$ & \cite{Kupyn_2019_ICCV} & $\Fen_1$ & $\Men_1$ & $I_1$ &  Static \\
\end{tabular}
\vspace{-2pt}
\caption{\textbf{FMOs in the wild,} 
captured by a mobile device and reconstructed by DeFMO and DeblurGAN-v2~\cite{Kupyn_2019_ICCV}. 
Our results are shown in 3 columns: estimated appearance $\Fen_1$, mask $\Men_1$, and the composed temporal super-resolution $I_1$.
Other methods, such as TbD~\cite{tbd} and Jin~\etal~\cite{Jin_2018_CVPR}, do not produce competitive results. 
}
\label{fig:wild}
\end{figure}

\renewcommand{\addimg}[1]{\includegraphics[width=0.045\linewidth]{#1}}

\newcommand{\makeRow}[2]{\multirow{2}{*}{\rotatebox[origin=c]{90}{#2}} & \addimg{#1_bgr} & \addimg{#1_sota18_0}  & \addimg{#1_deblurgan_hr} & \addimg{#1_tbdm0}&\addimg{#1_tbdfm0}&\addimg{#1_tbd0} & \addimg{#1_tbd3dm0}&\addimg{#1_tbd3dfm0}&\addimg{#1_tbd3d0} &  \addimg{#1_estm0}&\addimg{#1_estfm0}&\addimg{#1_est0} & \addimg{#1_hs0} & \addimg{#1_tbd3dom0}&\addimg{#1_tbd3dofm0}&\addimg{#1_tbd3do0} & \raisebox{1em}{\multirow{2}{1em}{$t_0$ $t_1$}} \\
 & \addimg{#1_im} &   \addimg{#1_sota18_7} & \addimg{#1_deblurgan_hr} & \addimg{#1_tbdm7}&\addimg{#1_tbdfm7}&\addimg{#1_tbd7} & \addimg{#1_tbd3dm7}&\addimg{#1_tbd3dfm7}&\addimg{#1_tbd3d7} &
 \addimg{#1_estm7}&\addimg{#1_estfm7}&\addimg{#1_est7} & \addimg{#1_hs7} & \addimg{#1_tbd3dom7}&\addimg{#1_tbd3dofm7}&\addimg{#1_tbd3do7} &  \\
\addlinespace[0.05em]
}
\newcommand{\ColSkip}{\hskip 0.2em}
\newcommand{\BlockSkip}{\hskip 2.9em}

\newcommand{\MakeArrow}{\multicolumn{3}{c}{\upbracefill \hspace*{0.2em}}}
\newcommand{\MakeArrowBlock}{\multicolumn{3}{c}{\upbracefill \hspace*{3.1em}}}

\begin{figure*}
\centering
\footnotesize
\setlength{\tabcolsep}{0.03em} 
\renewcommand{\arraystretch}{0.15} 

\begin{tabular}{@{}cc@{\ColSkip}c@{\ColSkip}c@{\ColSkip \ColSkip \BlockSkip}ccc@{\BlockSkip}ccc@{\BlockSkip}ccc@{\BlockSkip}c@{\BlockSkip}cccc@{}}

\makeRow{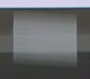}{box}
\makeRow{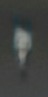} {key}
\makeRow{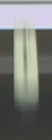}{marker}
\makeRow{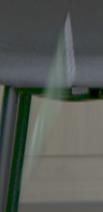}{pen}

\addlinespace[0.2em]
 & & & & \MakeArrowBlock  & \MakeArrowBlock  & \MakeArrowBlock & &  \MakeArrow & \\
\addlinespace[0.4em]

 & $B$ / $I$ & \cite{Jin_2018_CVPR} & \cite{Kupyn_2019_ICCV} & \multicolumn{2}{r}{TbD~\cite{tbd}} & & \multicolumn{3}{c}{TbD-3D~\cite{tbd3d}} &  \multicolumn{3}{c}{DeFMO (ours)} & GT & \multicolumn{3}{c}{TbD-3D-Oracle} & \\
\addlinespace[0.4em]

\end{tabular}
\caption{\textbf{Comparison on the Falling Objects dataset~\cite{kotera2020}} with the state-of-the-art methods: Jin~\etal~\cite{Jin_2018_CVPR}, DeblurGAN-v2~\cite{Kupyn_2019_ICCV}, TbD~\cite{tbd}, TbD-3D~\cite{tbd3d} and TbD-3D-Oracle that uses GT trajectories. For each method (except \cite{Jin_2018_CVPR,Kupyn_2019_ICCV} only showing deblurred results), we show from left to right: estimated mask, estimated sharp appearance, and temporal super-resolution frames for $t=0$ (top) and $t=1$ (bottom). 
}
\label{fig:results}
\vspace*{0.2em}
\end{figure*}


\boldparagraph{Ablation study} 
in Table~\ref{tbl:ablation} 
validated that the self-supervised losses (\ie all except for the supervised appearance reconstruction loss $\La_\Lfm$) have a positive impact on the convergence and generalization of the overall model.
Neglecting the sharpness loss $\La_\Ls$ generates more blurry object boundaries, which preserves the reconstruction quality but makes the trajectory less precise.
Training with only the appearance reconstruction loss leads to overfitting to the training set.
At the other extreme, training in fully self-supervised fashion 
completely fails (Table~\ref{tbl:ablation}, row 4).
We observed that the main problem was in identifying the object of interest and balancing the importance of the background motion. 
The combination of supervised and self-supervised losses shows the best performance.

\boldparagraph{Discussion}
None of the objects in the test datasets are present in ShapeNet but are still successfully reconstructed by our method.
DeFMO can also reconstruct a deforming object (Fig.~\ref{fig:results_full}, aerobie), even though only rigid bodies were used during training.
The method is able to deblur other dynamic objects simultaneously if their motion is similar to the object of interest,~\eg the volleyball in Fig.~\ref{fig:teaser}. 

\boldparagraph{Limitations}
When the object's appearance is similar to the background color, the problem becomes severely ill-posed.
For instance, the black tip of the marker in Fig.~\ref{fig:results} is not reconstructed, as the object was moving in front of the black background, and both reconstructions with and without the tip correctly lead to almost the same input image.

The method is not designed for objects made of transparent materials, \eg bottle, glass. The two sources of transparency, background-foreground mixing due to fast motion and the transparent material, are too difficult to distinguish. 


\boldparagraph{Applications}
of the proposed method include temporal super-resolution (Fig.~\ref{fig:results_full}). 
It can be used in fields such as astronomy to reconstruct the appearance of fast asteroids or data compression to decrease the frame rate at which a video is stored and then recover it back with DeFMO. 
Other applications are ball detection in sports and estimation of its speed or full 3D reconstruction of a highly blurred object by applying shape-from-silhouettes~\cite{visualhull}.
In combination with a standard tracker, DeFMO can track objects that are FMOs during parts of a video and not blurry in other parts.

DeFMO also works on sequences recorded in real-life settings, as on a mobile device. We captured videos of hand-thrown objects with a standard frame rate of 30~fps. Example of their reconstruction is in Fig.~\ref{fig:wild} (shoes) and in Fig.~\ref{fig:teaser} (lighter). FMO reconstruction of objects from YouTube videos is in Fig.~\ref{fig:teaser} (mic, cap).
More examples and videos are available in the supplementary material.


\section{Conclusion}
We proposed a novel generative model for disentangling and deblurring of fast moving objects.
Training on a complex synthetic dataset with a carefully designed loss function incorporating prior knowledge of the problem scales well to real-world data.
Experimental results show that the proposed model can handle fast moving objects with complex shapes and significant appearance changes within one video frame.
DeFMO sets a new state of the art as it outperforms all previous methods on multiple datasets.
Temporal super-resolution is among the possible applications.

\blfootnote{\textbf{Acknowledgements.} This work was supported by a Google Focused Research Award, the Czech Science Foundation grant GA18-05360S and Innosuisse grant No. 34475.1 IP-ICT.}

\newcommand{\addimCar}[1]{\includegraphics[width=0.23\linewidth]{imgs/white_car/races_0034_#1.png}}
\newcommand{\Write}[1]{\raisebox{0.5em}{\rotatebox[origin=lt]{0}{\scriptsize #1}}}

\begin{figure}
\centering
\small
\setlength{\tabcolsep}{0.02em} 
\renewcommand{\arraystretch}{0.15} 

\begin{tabular}{@{}cccc@{}c@{}}
    
\raisebox{1.0em}{\multirow{3}{*}{\begin{tabular}{c}\addimCar{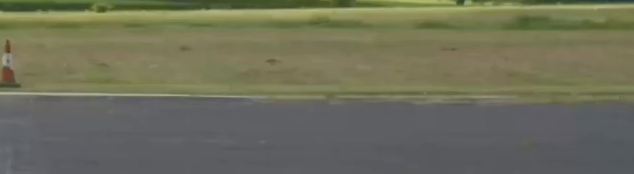} \\[0.5em]  \addimCar{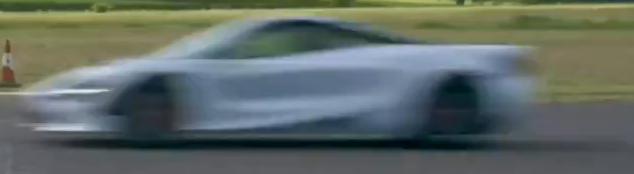} \\ \end{tabular}}} & \addimCar{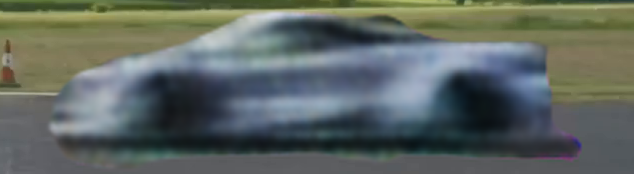} & \addimCar{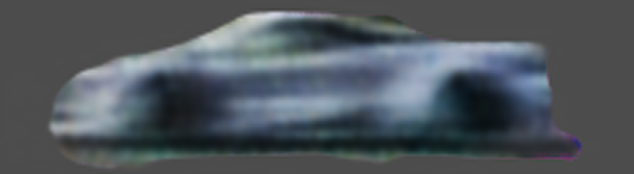}  & \addimCar{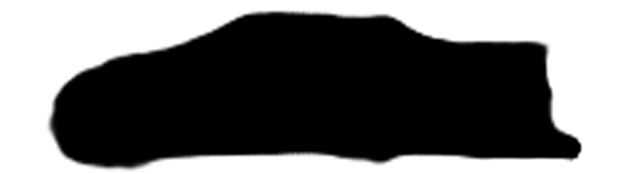} & \Write{0} \\
  & \addimCar{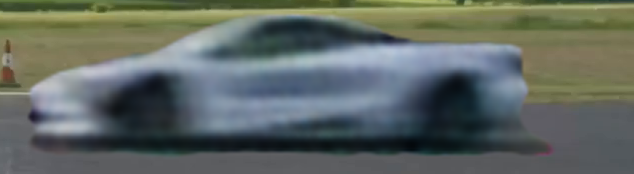} & \addimCar{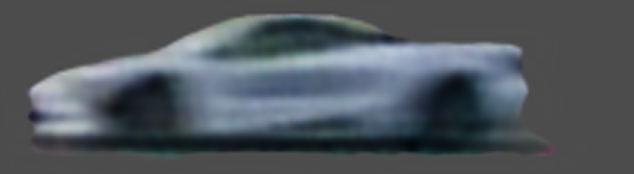}  & \addimCar{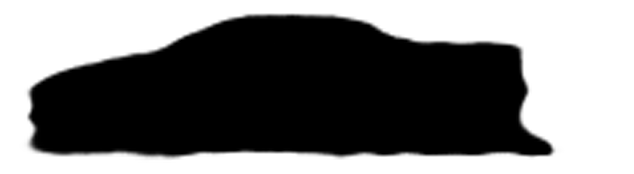} & \Write{0.5} \\
  & \addimCar{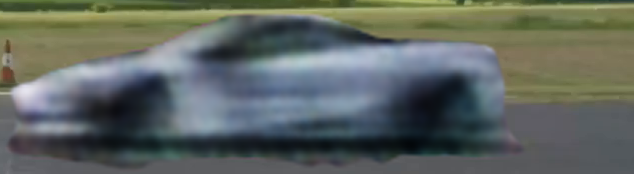} & \addimCar{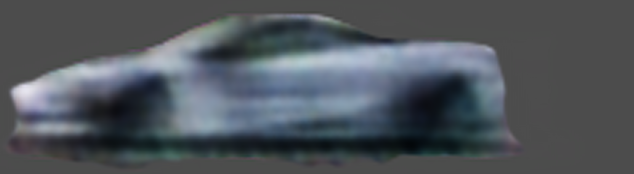}  & \addimCar{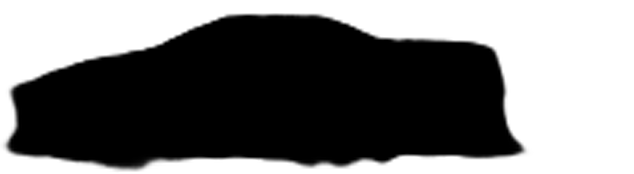} & \Write{1} \\
 Input ($B,I$) & $I_t$ & $F_t$ & $M_t$ & $t$ \\

\end{tabular}

\caption{\textbf{Results on a fast moving car}, available on \href{https://youtu.be/deAmxhcokd8?t=387}{YouTube}.}
\vspace*{-0.5em}
\label{fig:car}
\end{figure}
\newcommand{\addimgFuture}[1]{\includegraphics[width=0.1\linewidth,trim=0em 7em 0em 0em, clip=true]{imgs/key/v_pen_GTgamma_0007_#1.png}}
\newcommand{\MakeArrowFuture}[1]{\multicolumn{#1}{c}{\upbracefill \hspace*{0.2em}}}

\begin{figure}
\centering
\small
\setlength{\tabcolsep}{0.02em} 
\renewcommand{\arraystretch}{0.35} 

\begin{tabular}{@{}cccccccccc@{}}

\addimgFuture{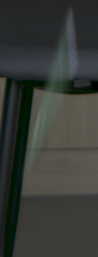} & \addimgFuture{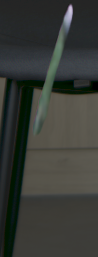} & \addimgFuture{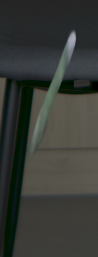} & \addimgFuture{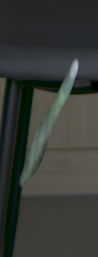}  & \addimgFuture{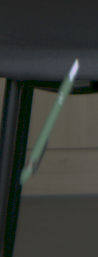}  & \addimgFuture{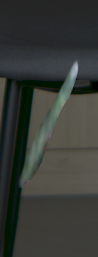} & \addimgFuture{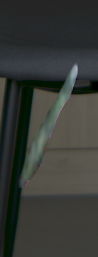} & \addimgFuture{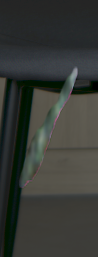} & \addimgFuture{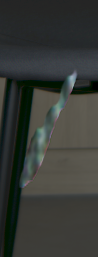} & \addimgFuture{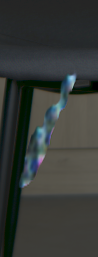}\\
$I$ & $I_0$ & $I_{0.5}$ & $I_1$  & GT$_{1}$ & $I_{1.05}$ & $I_{1.1}$ & $I_{1.2}$ & $I_{1.3}$ & $I_{1.4}$ \\
 & \MakeArrowFuture{3} & & \MakeArrowFuture{5} \\[0.1em]
 & \multicolumn{3}{c}{Standard DeFMO} & & \multicolumn{5}{c}{Future prediction} \\
 
\end{tabular}

\caption{\textbf{Estimated renderings} outside of the $[0,1]$ time range.}
\vspace*{-0.9em}
\label{fig:future}
\end{figure}
\renewcommand{\addimg}[1]{\includegraphics[width=0.057\linewidth]{#1}}

\begin{figure*}[h]
\centering
\footnotesize
\setlength{\tabcolsep}{0.01em} 
\renewcommand{\arraystretch}{0.1} 

\begin{tabular}{@{}cc@{\hskip 0.2em}ccccccccccccccccc@{}}

\makeRowFull{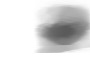}{football}
\makeRowFull{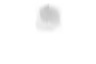}{key}

\addlinespace[0.2em]
\multicolumn{3}{c}{Inputs ($B,I$)} & 
\multicolumn{8}{c}{Temporal super-resolution} &
\multicolumn{8}{c}{Masks $M_t$} \\[0.5em]

\end{tabular}
\caption{\textbf{Temporal super-resolution on selected sequences from test datasets.} 
We compare to the TbD-3D-Oracle~\cite{tbd3d} with the manually provided ground truth trajectory from a high-speed footage (GT). Ground truth masks are computed as a difference image between the GT sub-frame and the background. The proposed DeFMO method estimates everything just from inputs on the left.}
\label{fig:results_full_add}
\end{figure*}
\renewcommand{\addimg}[1]{\includegraphics[width=0.043\linewidth]{#1}}

\begin{figure*}[h]
\centering
\footnotesize
\setlength{\tabcolsep}{0.03em} 
\renewcommand{\arraystretch}{0.15} 

\begin{tabular}{@{}cc@{\ColSkip}c@{\ColSkip}c@{\ColSkip \ColSkip \BlockSkip}ccc@{\BlockSkip}ccc@{\BlockSkip}ccc@{\BlockSkip}c@{\BlockSkip}cccc@{}}

\makeRow{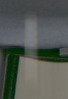}{cell}
\makeRow{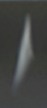}{eraser}

\addlinespace[0.2em]
 & & & & \MakeArrowBlock  & \MakeArrowBlock  & \MakeArrowBlock & &  \MakeArrow & \\
\addlinespace[0.4em]

 & $B$ / $I$ & \cite{Jin_2018_CVPR} & \cite{Kupyn_2019_ICCV} & \multicolumn{2}{r}{TbD~\cite{tbd}} & & \multicolumn{3}{c}{TbD-3D~\cite{tbd3d}} &  \multicolumn{3}{c}{DeFMO (ours)} & GT & \multicolumn{3}{c}{TbD-3D-Oracle} & \\
\addlinespace[0.2em]

\end{tabular}
\caption{\textbf{Comparison on the Falling Objects dataset~\cite{kotera2020}} with the state-of-the-art methods: Jin~\etal~\cite{Jin_2018_CVPR}, DeblurGAN-v2~\cite{Kupyn_2019_ICCV}, TbD~\cite{tbd}, TbD-3D~\cite{tbd3d} and TbD-3D-Oracle that uses GT trajectories. For each method (except \cite{Jin_2018_CVPR,Kupyn_2019_ICCV} only showing deblurred results), we show from left to right: estimated mask, estimated sharp appearance, and temporal super-resolution frames for $t=0$ (top) and $t=1$ (bottom). 
}
\label{fig:results_additional}
\end{figure*}

\renewcommand{\addimg}[1]{\includegraphics[width=0.09\textwidth]{#1}}

\begin{figure}
\centering
\footnotesize
\setlength{\tabcolsep}{0.03em} 
\renewcommand{\arraystretch}{0.15} 

\begin{tabular}{@{}c@{\hskip 0.5em}|@{\hskip 0.5em}ccccc@{}}

\makeRowSyn{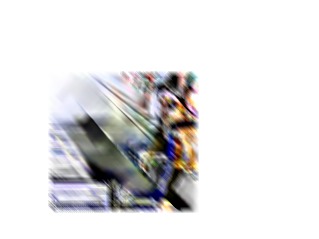} \\[0.1em]
 Inputs ($B,I$) & & $F_{0}$ & $F_{0.25}$ & $F_{0.5}$ & $F_{0.75}$ \\
 
\end{tabular}
\vspace{-0.3em}
\caption{\textbf{FMO reconstruction on a validation image from the synthetic dataset.} We compare to the TbD-3D-Oracle~\cite{tbd3d} method (with ground truth trajectory). 
Flying mug is synthetically superimposed.
}
\label{fig:results_syn_additional}
\end{figure}

{\small
\bibliographystyle{ieee_fullname}
\bibliography{egbib}
}

\end{document}